\pdfoutput=1

\documentclass[11pt]{article}

\usepackage[final]{acl}

\usepackage{times}
\usepackage{latexsym}

\usepackage[T1]{fontenc}

\usepackage[utf8]{inputenc}

\usepackage{microtype}

\usepackage{inconsolata}

\usepackage{graphicx}
\usepackage{multirow}
\usepackage{colortbl}
\usepackage{amsmath}
\usepackage{amssymb}
\usepackage{enumitem}
\usepackage{booktabs}
\usepackage{arydshln}
\usepackage[most]{tcolorbox}

\title{LLMs for Bayesian Optimization in Scientific Domains: Are We There Yet?}

\author{
    \textbf{Rushil Gupta}\textsuperscript{1,2},
    \textbf{Jason Hartford}\textsuperscript{3,4},
    \textbf{Bang Liu}\textsuperscript{1,2,5}\thanks{Corresponding author. Contact via email: bang.liu@umontreal.ca}
    \\[0.5em]
    \textsuperscript{1}DIRO, Universit\'e de Montr\'eal \& Institut Courtois \quad
    \textsuperscript{2}Mila - Quebec AI Institute \quad \\
    \textsuperscript{3}The University of Manchester\quad
    \textsuperscript{4}Valence Labs \quad
    \textsuperscript{5}Canada CIFAR AI Chair
}

\begin{document}
\maketitle
\begin{abstract}
Large language models (LLMs) have recently been proposed as general-purpose agents for experimental design, with claims that they can perform in-context experimental design. We evaluate this hypothesis using both open- and closed-source instruction-tuned LLMs applied to genetic perturbation and molecular property discovery tasks. We find that LLM-based agents show no sensitivity to experimental feedback: replacing true outcomes with randomly permuted labels has no impact on performance. Across benchmarks, classical methods such as linear bandits and Gaussian process optimization consistently outperform LLM agents. We further propose a simple hybrid method, \textit{LLM-guided Nearest Neighbour (LLMNN)} sampling, that combines LLM prior knowledge with nearest-neighbor sampling to guide the design of experiments. LLMNN achieves competitive or superior performance across domains without requiring significant in-context adaptation. These results suggest that current open- and closed-source LLMs do not perform in-context experimental design in practice and highlight the need for hybrid frameworks that decouple prior-based reasoning from batch acquisition with updated posteriors. 
\end{abstract}

\section{Introduction}
Experimental design enables data-efficient scientific discovery \cite{shields2021bayesian,ueno2016combo,lei2021bayesian, huan2024optimal, jain2023gflownets}. In domains such as biology and chemistry each experiment can be costly or time-consuming so we need methods that select the most informative candidates under a constrained budget. Classical approaches typically adopt a Bayesian framework \citep{Lindley56, Houlsby2011}: they construct a surrogate model of the response function, update this model with new data, and select future experiments by optimizing an acquisition function such as expected improvement or information gain \citep{gal2017deep, kirsch2019batchbald}. 

Recent work has proposed using LLMs to replace these components, offering a unified interface that can incorporate prior knowledge, reason over experimental history, and directly select candidates via prompting. Examples include BioDiscoveryAgent (BDA) \citep{roohani2024biodiscoveryagentaiagentdesigning} for genetic perturbation design and LLAMBO \citep{liu2024largelanguagemodelsenhance} for hyperparameter optimization. These systems are built on proprietary LLMs (e.g., Claude 3.5 Sonnet \citep{claudeAnthropic} or  OpenAI's GPT-3.5 \citep{gpt35} and o1 \cite{o1}) and design experiments by iteratively prompting the LLM with prior experimental outcomes, and rely on in-context learning to guide future experiment selection. 

In domains that are well-studied, the priors from the LLMs offer clear benefits over the traditional methods for the first round of experimentation because the LLMs can leverage information from the pretraining corpus to select actions. However, it is less clear whether in-context learning alone (without finetuning) leads to good action selection. 
This paper investigates whether such LLM-based approaches, when implemented with open-source models and without external tool use, can perform effective experimental design. Specifically, we ask, \textbf{do off-the-shelf, instruction-tuned LLMs exhibit strong in-context experiment design abilities when prompted with experimental history?} 

We address this question using the BioDiscoveryAgent pipeline implemented with three publicly available LLMs—\texttt{LLaMA-3.1-8B-Instruct} \citep{grattafiori2024llama3herdmodels}, \texttt{Qwen-2-7B-Instruct} \citep{yang2024qwen2technicalreport}, and \texttt{Qwen-2.5-14B-Instruct} \citep{qwen2.5}— and two closed-source LLMs— \texttt{Claude 4 Sonnet} \citep{claudeAnthropic4} and \texttt{GPT 4o-mini} \citep{4o-mini}—applied to experimental design tasks in two domains: single-gene perturbation and molecular property prediction. To evaluate in-context learning behavior, we conduct ablation studies that compare the standard BioDiscoveryAgent to a variant receiving randomly permuted feedback (BDA-Rand), removing any correspondence between past actions and outcomes. Our experimental analysis reveals that:
\begin{enumerate}
    \item \textbf{LLMs are insensitive to feedback.} Across all datasets and models (including Claude Sonnet 3.5), BDA and BDA-Rand perform comparably. In some cases, the BDA-Rand even performs slightly better. This suggests that the tested LLMs’ selection behavior is determined primarily by prior knowledge rather than adaptation to feedback.
    \item \textbf{LLMs underperform classical baselines.} Linear UCB and Gaussian process-based Bayesian optimization methods, given access to the same candidate embeddings, consistently outperform BDA on both domains.
    \item \textbf{A simple heuristic performs better search.} We introduce \textbf{LLM-guided Nearest Neighbour} (LLMNN), a method that prompts the LLM to propose seed candidates and then selects batches via nearest-neighbor expansion in embedding space. LLMNN also receives in-context feedback but outperforms BDA and matches or exceeds classical methods across benchmarks.
\end{enumerate}

In summary, we make following contributions,
\begin{itemize}
    \item We provide a diagnostic evaluation of in-context experimental design using randomized-feedback ablations.
    \item We benchmark open-source LLMs against Bayesian optimization and bandit-based methods across five gene perturbation and three molecular property prediction datasets.
    \item We show that open-source and proprietary instruction-tuned LLMs do not leverage experimental feedback to perform in-context experimental design in practical experimental design tasks.
    \item We introduce a simple hybrid method, LLMNN, which leverages LLM prior knowledge for exploration and embedding-based nearest-neighbor sampling for exploitation.
\end{itemize}

These results indicate that LLMs, while encoding valuable domain priors, still need explicit mechanisms that enable posterior updating and selection for efficient experimental design. Hybrid approaches that combine LLM priors with classical exploration strategies offer a promising direction for scalable, general-purpose experiment design.

\section{Problem Statement}
\label{sec:prob_stat}
Consider the following real-world experimental settings across two domains:

\textbf{Single Gene Perturbation.} Let's say we have the ability to knock out a single gene at a time from the human cell to induce a certain desired phenotypic response. Correspondingly, for every knockout, we are able to measure the phenotypic response value of perturbing the gene. However, exhaustive trials with the whole set of protein-coding genes ($> 19000$) are infeasible due to both time and budget constraints. Thus, \textit{how do we design perturbation experiments to identify genes that produce the phenotype under tight budget constraints and a sequential experiment protocol?}

\textbf{Chemical Properties of Molecules.} Similarly, say given a large library of molecules, we seek to identify the molecules that exhibit high values for a given property. For every molecule tried, we can obtain its property value. \textit{How do we effectively screen the library to identify top-performing molecules for the given property, under similar constraints as above?}\\

More formally, both these settings are examples of closed-loop experiment design, which is our center of focus in this work. At any experimental round, the agent determines which candidates to investigate next, given the results from all the prior experiments. The process continues for $N$ number of rounds, depending on the experimental budget available, with the objective of detecting a maximum number of hit candidates. 

Specifically, let $\mathcal{C}$ be the set of all candidates and $f: \mathcal{C} \rightarrow \mathbb{R}$ be the function that maps each candidate to a real-valued measurement. The task is inherently sequential, spread over $N$ rounds, where each round $i$ is defined as the trial of $\{c_1, c_2, \dots, c_B\} \subseteq \mathcal{C}$ candidates independently. The goal of the experiment is to identify the candidates with measurement value greater than some threshold $\tau$, i.e., $f(c) > \tau$ for $c \in \mathcal{C}$. These are termed as \textit{hits} for that measurement function.

Further, let $C_i$ be the set of candidates selected by the agent at any round $i$ and let $Z_i$ be the set comprising of entire experiment history, i.e. $Z_i = \{(c, f(c)) \mid c \in \bigcup\limits_{t=1}^{i-1} C_t\}$. Given $Z_i$, the agent is then expected to guide the formation of $C_i$.

At the end of $N$ rounds, let $\mathcal{C}_a = \bigcup\limits_{t=1}^{N} C_t^+$ , where
$C_t^+ = \{c \in C_t : f(c) > \tau \}$ represents the hits identified in round $t$. Thus, $\mathcal{C}_a$ represents the cumulative set of hits across all rounds up to $t$. Similarly, let $\mathcal{C}_{gt}$ be the set of all true hits for the measurement function $f$, $\mathcal{C}_{gt} = \{c \in \mathcal{C} : f(c) > \tau \}$.

We typically fix the $N = 5$ and $B = 128$ for all our experiments unless stated otherwise. Additionally, the $\tau$ for the molecular property task is set to be at the 90th percentile of the property value.

\section{Related Work}
\textbf{BO for Scientific Applications:} Bayesian Optimization in scientific domains has focused on a variety of applications like drug discovery \citep{griffiths2020constrained, korovina2020chembo}, biological experiment design \citep{roohani2024biodiscoveryagentaiagentdesigning, lyle2023discobax, king2004functional}, and chemical/molecular tasks \citep{kristiadi2024sober, fromer2024batchedbayesianoptimizationcorrelated, rankovic2023bochemian, shields2021bayesian}. Simultaneously, there have also been works that advance autonomous AI-driven closed-loop experiment design \citep{m2024augmenting, tom2024self, boiko2023autonomous}. 

In this work, we study how LMs can be used to incorporate prior information into the search process in the context of two real-world domains: \textit{single gene perturbations} and \textit{chemical property optimization}. Both tasks possess an extensive candidate space and complex relationships between the candidate and its associated measurement. However, the genetic domain has a finite candidate space of genes in the human genome, whereas the space of all molecule is far larger (up to $10^{60}$ small molecules \citep{bohacek1996art}) making it harder to have priors over any particular molecule.\\

\noindent \textbf{LLMs for Bayesian Optimization:} 
A series of recent works have explored the use of LLM embeddings in different ways, like for general-purpose regression \citep{nguyen2024predictingstringslanguagemodel}, to improve surrogate modeling \citep{ nguyen2024lico}, and to augment the traditional methods like Gaussian Processes \citep{Hartford2020, ramos2023bayesian}. We focus on using LLMs as the backend for an agent that interacts with the laboratory feedback and performs closed-loop experiment design. 

The two closest related prior works to our study are BioDiscoveryAgent \citep{roohani2024biodiscoveryagentaiagentdesigning} and LLAMBO \citep{liu2024largelanguagemodelsenhance}. LLAMBO is based on GPT-3.5 and primarily focuses on hyperparameter-tuning tasks, with LLM performing end-to-end Bayesian Optimization via suitable prompting. In a similar spirit, BioDiscoveryAgent proposes an LLM agent based on Claude 3.5 Sonnet and augmented with external tools that performs closed-loop experiment design of genetic perturbations. Both these approaches leverage proprietary LLMs and heavily rely on in-context learning to design experiments conditional on the experimental history. BioDiscoveryAgent, in particular, prompts the LLM to generate the entire batch of candidates to try next. In this work, we take a critical view of these methods, particularly BioDiscoveryAgent (BDA), and resort to instruction-tuned open-source models like Llama3.1-8B \citep{grattafiori2024llama3herdmodels} and Qwen2-7B \citep{yang2024qwen2technicalreport} to further examine it. While there exists evidence that transformers have been shown to perform amortized Bayesian inference \citep{müller2024transformersbayesianinference} when appropriately trained, it is not obvious that the ability arises from the next-token prediction and post-training objectives, and hence we specifically test the open-source BDA for the ability to select experiments. Eventually, we take a departure from a purely LLM-based approach and explore the synergy between LLMs and classical methods in our proposed hybrid approach, LLMNN, that achieves significantly superior performance.

\section{Tasks and Datasets}
\begin{table*}[htb]
\centering
\begin{tabular}{cccccc}
\toprule
\textbf{Method} & \textbf{IL2} & \textbf{IFNG} & \textbf{Carnevale} & \textbf{Sanchez} & \textbf{Sanchez Down}\\
\midrule
Ground truth ($\mid \mathcal{C}_{gt}\mid$) & 654 & 920 & 943 & 924 & 924\\
\midrule
\multicolumn{6}{c}{\textbf{Llama-3.1-8B backbone}}\\
BDA & 39.4 & 44 & 32.4 & 23.2 & 43.6\\
BDA-Rand & 37 & 51 & 31.6 & 30.8 & 45\\
\midrule
\multicolumn{6}{c}{\textbf{Qwen-2-7B backbone}}\\
BDA & 33.2 & 26.2 & 27.2 & 26.4 & 26\\
BDA-Rand & 29 & 32.4 & 29 & 24.2 & 35.4\\
\midrule
\multicolumn{6}{c}{\textbf{Claude 3.5 Sonnet backbone}}\\
BDA (Reported Numbers) & 68.01 & 87.4 & 39.6 & 60.72 & N/A\\
BDA (Replicated) & 59.4 & 78.8 & 43.8 & 31.6 & 51.8\\
BDA-Rand & 57.6 & 79.4 & 42 & 33.8 & 57.6\\
\bottomrule
\end{tabular}
\caption{Cumulative number of hits secured by each method on the gene perturbation datasets. The values are averaged over 5 runs. Note that we experiment with the No-Tool version of BioDiscoveryAgent. BDA-Rand is the BioDiscoveryAgent baseline provided with random feedback after each round of experimentation. It can be concluded that LLMs trained on next token prediction and RLHF fail to perform in-context experimental design.}

\label{tab:in-context-bayesian}
\end{table*}

\begin{table*}[htb]
\centering
\begin{tabular}{cccccc}
\toprule
\textbf{Method} & \textbf{IL2} & \textbf{IFNG} & \textbf{Carnevale} & \textbf{Sanchez} & \textbf{Sanchez Down}\\
Ground truth ($\mid \mathcal{C}_{gt}\mid$) & 654 & 920 & 943 & 924 & 924\\
\midrule
\multicolumn{6}{c}{\textbf{Llama-3.1-8B backbone}}\\
Linear UCB & 35 & \textbf{72} & \textbf{38} & \textbf{39} & 42\\
GP & \textbf{147.8} & 23 & 22.2 & 27.6 & 30\\
BDA & 39.4 & 44 & 32.4 & 23.2 & \textbf{43.6}\\
\midrule
\multicolumn{6}{c}{\textbf{Qwen-2-7B backbone}}\\
Linear UCB & 93 & \textbf{74} & \textbf{31} & \textbf{31} & \textbf{41}\\
GP & \textbf{147.8} & 23 & 22.2 & 27.6 & 30\\
BDA & 33.2 & 26.2 & 27.2 & 26.4 & 26\\
\bottomrule
\end{tabular}
\caption{Cumulative number of hits secured by each method on the gene perturbation datasets. The values are averaged over 5 runs. The max values in each column for each LLM backbone have been bolded. Note that we experiment with the No-Tool version of BioDiscoveryAgent. Further, all the statistical methods use the embeddings corresponding to the LLM in the backbone of BDA to ensure the same amount of knowledge. It can be observed clearly that principled statistical approaches outperform the BioDiscoveryAgent baseline on most datasets and across both families of open-source LLMs.}
\label{tab:bda-std-baselines}
\end{table*}

\label{sec:tasks}
We perform experiments on two scientific domains in this work. In particular, we work with \emph{single gene perturbations} and \emph{chemical properties of molecules}. We simulate running experiments on a set of candidates by retrieving the corresponding measurement value from a tabular dataset. However, in real-world settings, the corresponding experiments would be conducted in a laboratory or an in-silico simulator. In the subsequent subsections, we describe the datasets used in the two domains.
\subsection{Single Gene Perturbations}
Section \ref{sec:prob_stat} explains the details of the task, such as the goal with experiment design, candidate space, and the associated measurements. In terms of datasets, we use the \textbf{IL2, IFNG, Carnevale, Sanchez, and Sanchez Down} datasets from \citet{roohani2024biodiscoveryagentaiagentdesigning}'s BioDiscoveryAgent as each of them covers a distinct biological process. For example, \textbf{IL2} \citep{schmidt2022crispr} is concerned with the phenotype of change in the production of Interleukin-2 (IL2), a cytokine involved in immune signaling. More details about each of the datasets are included in the appendix \ref{sec:gene-perturb-datasets}. Each dataset contains measurements for over $18000$ genes, i.e., $\mid\mathcal{C}\mid > 18000$, each knocked down in a distinct cell.

\subsection{Chemical Properties of Molecules}
\label{sec:chem-dataset}
For the chemical properties task, we focus on the following three molecular datasets: \textbf{ESOL} (water solubility), \textbf{FreeSolv} (hydration free energy in water), and \textbf{Ion. E.} (ionization energy of the molecules). Note that $\mid\mathcal{C}\mid = 1128, 642$ and $11565$ respectively for each of the aforementioned datasets. For more details, see appendix \ref{sec:chem-datasets-app}

\section{Is Naively Prompting LLM Enough for Experiment Design?}
Experimental design is inherently a sequential process where at each round, one should select experiments on the basis of both prior information \emph{and} the outcomes of previous experiments. Through the pretraining objective, LLMs have extensive prior information which enables strong performance in experimental design settings with relatively few rounds of experiments (i.e. where the guidance of the prior matters most). For example, LLAMBO \citep{liu2024largelanguagemodelsenhance} focuses on tasks related to hyperparameter-tuning, and BioDiscoveryAgent \citep{roohani2024biodiscoveryagentaiagentdesigning} demonstrates strong performance on genetic perturbation experiment design. Despite minor differences, there are major similarities between these approaches. Both works pass on the experiment history and corresponding observations by simply appending them within the prompt itself with the hope that the LLM can leverage its in-context abilities to incorporate this feedback and adapt its subsequent predictions. We test the extent to which BioDiscoveryAgent is actually using this information by breaking the relationship between candidates, $c_i$, and their associated outcomes, $f(c_i)$. In particular, we randomly pair each $c_i$ with some other outcome, $f(c_j)$, which breaks their joint dependency, while keeping their respective marginal distributions fixed.

\begin{figure}[ht]
\includegraphics[width=\columnwidth]{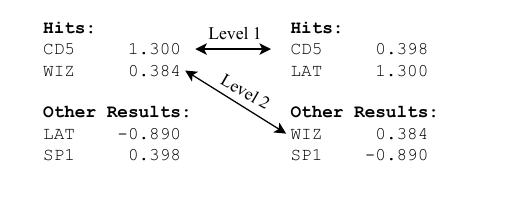}
  \caption{Illustration of the 2 levels of random permutation in the experimental feedback to the LLM on the genetic domain. Level 1 modifies the measurement value, whereas level 2 modifies whether a candidate is a hit or not. Note that it is possible that a particular gene has both level 1 and 2 modifications.
  }
  \label{fig:rand-illus}
\end{figure}

\textbf{Method.} For our investigation, we work with the BioDiscoveryAgent (BDA) pipeline on a small open-source LLM backbone without any external tools. In the first set of experiments, we evaluate the ability of the LLMs to leverage experimental feedback. We compare BDA with BDA-Rand, which is the same as BDA but receives randomized feedback instead of true feedback. We perform 2 levels of randomization, level 1 being random measurement values and level 2 being randomness in hit vs not-hit feedback. Figure \ref{fig:rand-illus} contains an illustration of the randomization procedure. Both methods are evaluated against the ground truth hits for each dataset. As a second step, we compare BDA with classical models like Linear UCB and Gaussian Process (GP) to evaluate whether the LLM has sufficient information to make strong selections. Both Linear UCB and the GP only condition on the residual stream embeddings of the LLM so can only outperform the LLM insofar as they can better use the experimental feedback.

\textbf{Results.} We investigated the in-context experimental design abilities of the LLMs, and Table \ref{tab:in-context-bayesian} shows the results for the same. The numbers in the table correspond to the cumulative number of hits obtained after 5 rounds of experiments with 128 perturbations in every round and are averaged over 5 runs. Across both LLMs and all 5 datasets, it is evident that passing random feedback does not hurt the performance of the framework at all. In fact, the performance remains nearly the same or improves slightly. To further substantiate our hypothesis, we also perform a similar experiment on Claude 3.5 Sonnet, a proprietary large-scale LLM, and observe that even Claude maintains nearly the same performance despite random feedback \footnote{We generate the numbers for BDA using the publicly available code as we could not replicate their original numbers, possibly due to LLM updates.}. The strong initially performance of the LLMs is therefore likely the result of theirs priors on ordering of genes and is not affected by the feedback of past experiments appended in its prompt.

\begin{figure*}[ht]
  \includegraphics[width=\linewidth]{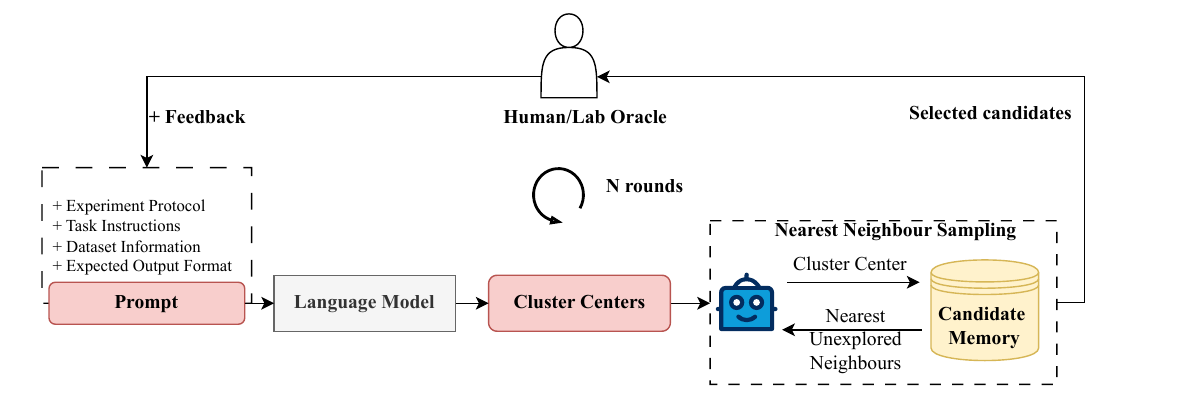}
  \caption{Schematic Workflow of the LLM-guided Nearest Neighbour (LLMNN) method. The red rounded boxes are the inputs and outputs of the Language Model. For ease, only cluster centers have been shown in the output. Please refer to Section \ref{sub:response-format} for detailed output specification and information about the components of the method. Refer to Appendix \ref{app:prompt-template} for the detailed prompt template and to Appendix \ref{sec:full-trace} for a full trace of the method.}
  \label{fig:diagram}
\end{figure*}

The above flaw in the LLM's capabilities motivated us to compare BDA with classical approaches. Table \ref{tab:bda-std-baselines} shows the cumulative number of hits obtained for BDA and classical baselines as described above. It can be seen clearly that across the 5 datasets, either LinearUCB or GP outperforms the BDA framework by a significant margin on both choices of LLM models. In contrast to the trend of Claude-based BDA, these results further highlight the lack of robustness of the framework's performance with respect to the base LLM.

\section{LLMNN: A Hybrid Experiment Design Method}
In light of the above observations, we propose \textbf{LLM} guided \textbf{N}earest \textbf{N}eighbour (LLMNN) framework, a simple greedy approach that prompts an LLM to guide the location of cluster centers and leverages nearest neighbour sampling to form the batch $B_i$ at any round $i$. The LLMNN framework is designed to leverage: 1.) Generalist LLMs’ intrinsic domain knowledge to guide the search in the vast candidate space, and 2) the inductive bias that similar genes/molecules have similar functionalities/properties. The schematic workflow of the proposed LLMNN framework is illustrated in Figure \ref{fig:diagram}. Below, we discuss the important components of the LLMNN framework, followed by a detailed workflow of the method.\\

\noindent\textbf{Candidate Memory.} stores the pool of candidates, each represented in the following format: \texttt{\{"name": <candidate name>, "score": <measurement value>, "explored": <bool>\}}. For molecules, the candidate name is the SMILES string and measurement is the property value, whereas for genes, the candidate name is the gene name in the HGNC nomenclature and measurement is the phenotypic response value. Each candidate is also associated with an embedding. The explored key tracks whether a candidate has already been selected, preventing repetition in the optimization process. The memory interacts with the LLMNN framework by receiving a query candidate name and returning the nearest unexplored neighbors based on distance in the embedding space.\\

\noindent\textbf{Candidate Embeddings.} Each candidate in the memory has an associated embedding. We experiment both with LLM embeddings for the candidates and also domain-specific embeddings like Achilles embeddings for genes \citep{tsherniak2017defining} and Molformer \citep{ross2022large} embeddings for the SMILES strings. The distance metric for Achilles embeddings is the cosine distance, whereas for all other embeddings, the L2-squared distance is used. More details about the embeddings are included in Appendix \ref{sec:embeds-details}\\

\noindent\textbf{Agent Response Format:} \label{sub:response-format} We borrow the agent response format from BioDiscoveryAgent \citep{roohani2024biodiscoveryagentaiagentdesigning}, and direct the LLM to structure its responses into several parts: \texttt{Reflection, Research Plan, Solution}, similar to \citet{huang2023mlagentbench}. Through the \texttt{Reflection} and \texttt{Research Plan} entries, the model is able to articulate its reasoning behind a particular prediction. \texttt{Solution} contains a list of $n_c$ cluster centers to sample around next. For our experiments, we fix $n_c = 5$ unless stated otherwise.\\

\begin{table*}[ht]
\centering
\begin{tabular}{cccccc}
\toprule
\textbf{Method} & \textbf{IL2} & \textbf{IFNG} & \textbf{Carnevale} & \textbf{Sanchez} & \textbf{Sanchez Down}\\
\midrule
Ground truth ($\mid \mathcal{C}_{gt}\mid$) & 654 & 920 & 943 & 924 & 924\\
Random & 23.8 & 26 & 35 & 32.2 & 34.6\\
Coreset & 9.2 & 31.4 & 29.4 & 33.8 & 30.2\\
Linear UCB & 110 & 98 & 51 & \textbf{75} & \textbf{87}\\
GP  & 139 & 24 & 20.4 & 24.6 & 33\\
BDA-GS (Claude 3.5 Sonnet) & 65.4 & 88.32 & 40.55 & 57.28 & N/A\\
\midrule
\multicolumn{6}{c}{\textbf{Llama-3.1-8B backbone}}\\
BDA-GS & 63.4 & 57.4 & 42.59 & 22.2 & 39.6\\
LLMNN NoExp & 137.8 & 86.3 & 58.9 & 42.5 & 42.3\\
LLMNN & 163.3 & 107.9 & 65.6 & 54.2 & 45.5\\
\midrule
\multicolumn{6}{c}{\textbf{Qwen-2-7B backbone}}\\
BDA-GS & 38.6 & 35.2 & 32.8 & 28.6 & 38\\
LLMNN NoExp & 146.4 & 59.1 & 44.5 & 62.8 & 46.4\\
LLMNN & 160.7 & 78 & 60.9 & 53.5 & 45.1\\
\midrule
\multicolumn{6}{c}{\textbf{Qwen-2.5-14B backbone}}\\
BDA-GS & 49 & 63.7 & 47 & 39.2 & 48.4 \\
LLMNN NoExp & 88.6 & 57.8 & 58.6  & 63.4 & 48.2 \\
LLMNN & 151.8 & 88 & 66.4 & 62.8 & 43 \\
\midrule
\multicolumn{6}{c}{\textbf{Claude 4 Sonnet backbone}}\\
BDA-GS & 65.2 & 83.6 & 38.6 & 45.8 & 60 \\
LLMNN NoExp & \textbf{179.4} & 95.2 & \textbf{68.2 } & 63.6 & 47.2\\
LLMNN & 159 & 89.8 & 62.4 & 59.2 & 45.6 \\
\midrule
\multicolumn{6}{c}{\textbf{GPT 4o-mini backbone}}\\
BDA-GS & 38.8 & 50.6 & 34 & 27 & 43.5\\
LLMNN NoExp & 117.4 & \textbf{108.2} & 55 & 64.4 & 41.2\\
LLMNN & 165 & 97 & 54.8 & 63.2 & 45.8 \\
\bottomrule
\end{tabular}
\caption{Cumulative hits over 5 rounds of experiments with 128 candidates in each experiment. The values in the table are averaged over 5 runs. The max values in each column have been bolded. Note that we use Achilles embeddings of genes for the gene search tool in the case of LLMNN and BDA-GS on Llama and Qwen backbones. The results indicate the strong performance of LLMNN, even surpassing the BDA based on Claude 3.5 Sonnet, which has access to a sophisticated gene search tool. Since the numbers for this BDA variant have been taken from Table 3 of the original paper, we do not have a score for Sanchez Down as it isn't included in the original text.} 
\label{tab:all-methods}
\end{table*}

\begin{table}[ht]
\centering
\begin{tabular}{cccc}
\toprule
\textbf{Method}  & \textbf{Ion. E.} & \textbf{ESOL} & \textbf{FreeSolv}\\
\midrule
G. truth ($\mid \mathcal{C}_{gt}\mid$) & 1156 & 112 & 64\\
Random & 77.8 & 29 & 14.4\\
Coreset & 151.6 & 57.6 & 20.4\\
Linear UCB & \textbf{283} & \textbf{76} & 39\\
GP  & 151.39 & 35 & 16.8 \\
\midrule
\multicolumn{4}{c}{\textbf{Llama-3.1-8B backbone}}\\
LLMNN NoExp & 152.8 & 23.4 & 12.2\\
LLMNN & 103.2 & 22.2 & 10.2\\
\midrule
\multicolumn{4}{c}{\textbf{Qwen-2-7B backbone}}\\
LLMNN NoExp & 147.2 & 24 & 25.8 \\
LLMNN & 134.8 & 36.2 & 23\\
\midrule
\multicolumn{4}{c}{\textbf{Qwen-2.5-14B backbone}}\\
LLMNN NoExp & 185.6 & 33.6 & 17.2 \\
LLMNN & 136.2 & 32 & 12.6\\
\midrule
\multicolumn{4}{c}{\textbf{Claude 4 Sonnet backbone}}\\
LLMNN NoExp & 173.3 & 60.6 & \textbf{43.6}\\
LLMNN & 189.6 & 63.8  & 38\\
\midrule
\multicolumn{4}{c}{\textbf{GPT 4o-mini backbone}}\\
LLMNN NoExp & 179.8 & 27.4 & 34.2\\
LLMNN & 119.8 & 31 & 29.8\\
\bottomrule
\end{tabular}
\caption{Cumulative hits over 5 rounds of experiments with $B = 128, 64,$ and $32$ candidates in each experiment and $n_c = 5,4$ and $4$, respectively, for Ion. E., ESOL and FreeSolv datasets. The values are averaged over 5 runs. Note that we use embeddings from MolFormer XL-10pct for molecule similarity. The results indicate that the NoExp variants of LLMNN perform competitively compared to classical baselines on 2/3 datasets.}
\label{tab:all-methods-mols}
\end{table}

\begin{table*}[htb]
\centering
\begin{tabular}{cccccc}
\toprule
\textbf{Method} & \textbf{IL2} & \textbf{IFNG} & \textbf{Carnevale} & \textbf{Sanchez} & \textbf{Sanchez Down}\\
\midrule
Random Centroids & 76 & 53.6 & 44.6 & 36.8 & 29.4 \\
LLMNN NoExp & \textbf{179.4} & \textbf{95.2} & \textbf{68.2 } & \textbf{63.6} & \textbf{47.2}\\
\bottomrule
\end{tabular}
\caption{Cumulative number of hits secured by the random centroids ablation variant and the best performing LLMNN NoExp method based on Claude 4 Sonnet on the gene perturbation datasets. The Achilles embeddings are used for gene similarity. The values are averaged over 5 runs.}
\label{tab:rand-centroid-gene}
\end{table*}

\begin{table}[htb]
\centering
\begin{tabular}{cccccc}
\toprule
\textbf{Method} & \textbf{Ion. E.} & \textbf{ESOL} & \textbf{FreeSolv}\\
\midrule
Rand. Centroids & 83.2 & 24 & 16 \\
LLMNN NoExp & \textbf{173.3} & \textbf{60.6} & \textbf{43.6}\\
\bottomrule
\end{tabular}
\caption{Cumulative hits secured by the random centroids and the best performing Claude 4 Sonnet-based LLMNN NoExp variant on the molecular datasets. Molformer embeddings are used for molecule similarity.}
\label{tab:rand-centroid-mols}
\end{table}

\begin{table}[htb]
\centering
\begin{tabular}{cccc}
\toprule
\textbf{Embeds}  & \textbf{Ion. E.} & \textbf{ESOL} & \textbf{FreeSolv}\\
\midrule
\multicolumn{4}{c}{\textbf{Claude 4 Sonnet backbone}}\\
Molformer & 189.6 & 63.8 & 38\\
OpenAI & 200.2 & 75.4  & 39.6\\
\midrule
\multicolumn{4}{c}{\textbf{GPT 4o-mini backbone}}\\
Molformer & 119.8 & 31 & 29.8\\
OpenAI & 169 & 46 & 46.2\\
\bottomrule
\end{tabular}
\caption{Cumulative hits by LLMNN method with different embeddings over 5 rounds of experiments with same experiment settings as Table \ref{tab:all-methods-mols}. The values in the table are averaged over 5 runs. OpenAI embeddings correspond to the OpenAI's text-embedding-3-large model.}
\label{tab:diff-embeds-mols}
\end{table}

\noindent\textbf{Workflow.} The exact flow of the LLMNN framework is defined as follows: \vspace{-2mm}
\begin{enumerate}[leftmargin=*]
    \item LLMNN framework takes experiment protocol, task, and candidate space details as the context in its prompt to generate the first set of cluster centers it seeks to explore. \vspace{-1mm}
    \item The output cluster centers are then used to retrieve candidates from the candidate memory.\vspace{-1mm}
    \item The selected candidates for the round are then sent to the human or a lab oracle to obtain their value of the measurement function $f$ and to determine if they were a hit.\vspace{-1mm}
    \item This hit vs not hit qualitative feedback, along with the measurement value, is appended to the prompt from Step 1 and passed to the LLM to generate the next set of cluster centers.\vspace{-1mm}
    \item Steps 2-4 are repeated for $N$-1 rounds. The total hits identified successfully at the end indicate the method's performance.
\end{enumerate}

\section{Does LLMNN do Any Better on Our Experiment Design Domains?}
We observed BioDiscoveryAgent struggled to outperform the traditional baselines (when provided with the same priors via access to embeddings) on the genetic perturbation datasets, and that the performance was largely the result of having better priors.
Given this, one would expect better performance from LLM-based approaches in settings with well-studied candidate spaces---e.g. genes in the human genome---compare with much larger spaces---e.g. all of chemical space.

LLMNN addresses the poor response to experimental feedback by
taking advantage of the fact that LLMs encode candidates into a common embedding space. This allows us to greedily construct experimental batches by performing nearest neighbor sampling in the neighborhood of previously observed hits. LLMNN requires minimal assumptions of LLM generating valid gene names as per HGNC nomenclature and valid SMILES strings, which is reasonable for modern-day LLMs that have been pretrained on an internet-scale of knowledge. Further, LLMNN is an attempt at exploiting the known inductive biases of the domain to achieve higher performance. 

\textbf{Method.} We experiment with two variants of LLMNN, one original with the \texttt{Research Plan} and \texttt{Reflection} outputs, and the other that contains purely the \texttt{Solution}, without any explanations. Both LLMNN variants have access to the gene similarity tool based on the Achilles \citep{tsherniak2017defining} embeddings. Against LLMNN, we include traditional baselines like Random, Coreset (pure diversity-based approach), LinearUCB, and Gaussian Process (GP). Apart from these methods, we also include three variants of BioDiscoveryAgent: 1) BDA, i.e., without any tools, 2) BDA-GS with access to the same gene similarity tool that LLMNN has access to, and 3) BDA-GS (Claude 3.5 Sonnet), which is the originally proposed approach in \citet{roohani2024biodiscoveryagentaiagentdesigning} that uses Claude 3.5 Sonnet LLM in the backbone and has access to more sophisticated gene similarity search tools like enrichment analysis on Reactome database \citep{gillespie2022reactome}. We do not include BDA in the molecules domain because BDA is not constrained to predict molecules that exist in the library and hence one needs a large number of retries to construct a batch. The tool for molecular similarity available to LLMNN is based on Molformer \citep{ross2022large} embeddings of SMILES strings.

\textbf{Results.} Table \ref{tab:all-methods} displays the performance comparison between different methods on the single gene perturbation domain. Note that the same experimental budget has been provided to all the methods, and the numbers are averaged over 5 runs. The table illustrates that the LLMNN method based on the Llama-3.1 backbone outperforms BDA based on both Llama and Claude 3.5 Sonnet backends on 5/5 and 3/4 datasets, respectively, by significant margins. It is worth noting that Llama-3.1 is just an 8 billion parameter model as compared to Claude 3.5, which has been trained on larger data with significantly more parameters. Further, LLMNN only had access to a basic gene similarity tool as compared to Claude 3.5 BDA, which had more sophisticated gene search tools. Similarly, Qwen-2-7B-based LLMNN outperforms the corresponding BDA on 5/5 datasets. Moreover, the LLMNN method continues to outperform the corresponding BDA while outperforming or closely matching classical baselines on all datasets with larger LLMs as backbones, especially Claude 4 Sonnet. Another interesting observation is that the traditional baselines still perform really well, surpassing Claude 3.5 BDA on 4/4 datasets and LLMNN as well on 2/5 datasets. One of the contributing factors to the performance of LLMNN is maintaining a memory that keeps track of which genes have already been explored. This ensures that similarity queries return unexplored neighbours at every query, in contrast to the BioDiscoveryAgent, which doesn't maintain this state and hence, would inevitably return the same set of genes always for the same query, irrespective of the experiment history.

On the molecular domain, Table \ref{tab:all-methods-mols} contains the numbers for cumulative hits for all methods averaged over 5 runs, given the same experimental budget. The numbers highlight that while Qwen2-7B-based LLMNN without explanations performs closely to the traditional baselines on 2/3 datasets, the latter still wins over the LLMNN approach across all the datasets, especially with the ESOL dataset. However, with larger LLMs like Claude 4 Sonnet, the LLMNN method outperforms classical methods on the FreeSolv dataset while significantly reducing the gap on the other 2 datasets. This underscores that LLMs with simple inductive biases are strong but still not enough alone to serve as experiment designers across different scientific domains. They need tight coupling with the more principled statistical methods that trade off exploration and exploitation, like linear UCB and GPs, to achieve higher performance.

Given the strong performance of the LLMNN method, we perform an ablation with LLM being replaced by a random centroid selector in the method. Our goal is to observe if LLM guidance plays any role in the superior performance of our method or the gains are purely due to nearest neighbour sampling in a strong embedding space. Tables \ref{tab:rand-centroid-gene} and \ref{tab:rand-centroid-mols} show the result comparing the ablation variant with the best-performing LLMNN NoExp variant, i.e., the one based on the Claude 4 Sonnet. We observe that the performance drops sharply, by over 50\% in some cases, when removing LLM guidance, underscoring its central role in deciding the centroids for nearest neighbour sampling.

Further, we also investigate if the gains from our method on molecular domains hold with a general-purpose LLM embedder like OpenAI's text-embedding-3-large model that isn't directly trained on SMILES strings. As shown in table \ref{tab:diff-embeds-mols}, LLMNN consistently improves performance, demonstrating the robustness of our approach across meaningful embedding spaces.

\section{Conclusion}
In this work, we set out to critically examine this hypothesis using instruction-tuned open- and closed-source LLMs. We particularly focused on BioDiscoveryAgent and experimented on two domains: single gene perturbations (5 datasets) and molecular property optimization (3 datasets).
When compared to classical methods like Linear UCB and Gaussian Process that used the embedding from the same LLM as BDA, the open-source BDA performed significantly worse. Furthermore, on deeper experiments with random feedback to the LLM's context, the open-source and Claude 3.5 Sonnet-based BDA still retained a similar average performance, showcasing that the LLMs trained on next token prediction and RLHF do not leverage experimental feedback in the design of their experiments.
Finally, we proposed an LLM-guided Nearest Neighbour framework (LLMNN), a simple combination of LLM and a classic nearest neighbour method. LLMNN outperforms BDA significantly on gene datasets and is applicable to molecular domains, and performs at par with the classical baselines, except on some molecular datasets. Overall, this work suggests that more work is needed to effectively incorporate experimental feedback into LLM-based experimental design pipelines. It introduces a plausible avenue for future research marked by the synergy of LLMs, classical methods, and domain-specific inductive biases.

\section*{Limitations and Future Works}
While we demonstrate superior performance with LLMNN on both gene perturbation and molecular domains, this study has several limitations. Firstly, LLMNN augments the LLM with a very simplistic nearest neighbour sampling that allocates an equal budget to each cluster to generate better predictions. However, more complex schemes can be explored to adaptively allocate more budget to centers that have a higher probability of detecting hits. For example, a probabilistic model like GP could be used to determine the hit likelihood of the predicted centers that can, in turn, be used to define budget allocation between the centers. 

Due to the policy of sampling around centers, the method is largely exploitative and hence sensitive to the choice of embeddings and hits identified in earlier rounds. More tightly-coupled integration of LLMs and classic exploration methods could be investigated to improve the robustness of the experiment design agent. Further, the inductive bias that \textit{similar} candidates have \textit{similar} properties is clearly not the best bias on molecular domains, as the classic exploration approaches maintain a strong performance as compared to LLMNN, suggesting the need to identify and encode more domain-specific and task-specific biases.

Lastly, another promising direction is to explore how external tools like literature search, enrichment analysis tools for genes, etc. be effectively augmented to the agent for better performance, as also highlighted by \citet{roohani2024biodiscoveryagentaiagentdesigning}.

\section*{Ethical Considerations}
This work deals with building hybrid AI agents that are powered by LLMs and classical exploration settings for guiding the design of scientific experiments. Since it involves the use of LLM, it is vulnerable to adversarial attacks where the agent might suggest experiments that lead to catastrophic outcomes in a real laboratory. Therefore, it is imperative for the users to have human scientists in the loop to screen out such experiment configurations. Not just post-hoc, the safety could be ensured at the start by suitable constraints on the candidate space.

One of the domains in this work is single gene perturbations, where the effect of knocking down a gene is measured in human cells. While these agents will help increase the robustness of the target discovery phase of the drug discovery pipeline, it should be taken into account that human cells often exhibit genetic variation from one population to another, thus limiting the transferability of the downstream insights to new groups. Thus, caution needs to be exercised in utilizing the insights drawn by this agent in the actual drug discovery pipeline.

\section*{Acknowledgments}
This work is supported by the Canada CIFAR AI Chair Program and the Canada NSERC Discovery Grant (RGPIN-2021-03115).

\bibliography{custom}

\appendix
\section{Prompt Templates}
\label{app:prompt-template}
In the following subsections, we provide the detailed prompt template used for both the genetic perturbation and molecular property domains.

\begin{table}[t]
\centering
\begin{tabular}{|p{0.2\columnwidth}|p{0.4\columnwidth}|p{0.3\columnwidth}|}
\hline
\rowcolor[gray]{0.9}
\textbf{Dataset} & \textbf{func desc} & \textbf{score desc}\\
\hline
IL2 & regulate the production of Interleukin-2 (IL-2) & log fold change in Interleukin-2 (IL-2) normalized read counts\\
\hline
IFNG & regulate the production of Interferon-gamma (IFNG) & log fold change in Interferon-gamma (IFNG) normalized read counts \\
\hline
Carnevale & upon being knocked out, would boost the efficacy of engineered T cells in the presence of an adenosine agonist that creates an immunosuppresive condition & change in T cell proliferation \\
\hline
Sanchez & when knocked out, either increase or decrease expression of endogenous tau protein levels in neurons & change in tau protein level compared to the non-targeting control, using a total tau antibody\\
\hline
Sanchez Down & when knocked out, decrease expression of endogenous tau protein levels in neurons & change in tau protein level compared to the non-targeting control, using a total tau antibody\\
\hline
\end{tabular}
\caption{func desc and score desc for the different gene perturbation datasets}
\label{tab:desc-for-genes}
\end{table}

\begin{table}[t]
\centering
\begin{tabular}{|p{0.2\columnwidth}|p{0.3\columnwidth}|p{0.3\columnwidth}|}
\hline
\rowcolor[gray]{0.9}
\textbf{Dataset} & \textbf{func desc} & \textbf{candidate space info}\\
\hline
Ion. E. & ionization energy (in eV) & The molecules in the library are composed of only C, H, N and O elements.\\
\hline
ESOL & solubility in water (log mol per litre) & The molecules in the library are small organic molecules. \\
\hline
FreeSolv & hydration free energy in water & The molecules in the library are small organic molecules. \\
\hline
\end{tabular}
\caption{func desc and candidate space info for the different molecular property datasets}
\label{tab:desc-for-mols}
\end{table}

\subsection{Single Gene Perturbation}
\fbox{\begin{minipage}{\linewidth}
\texttt{
\textbf{SYSTEM PROMPT:}\\
You are a biomedicine expert who will assist me on problems in drug discovery. I am planning to run a CRISPR screen to identify genes that \textbf{\{func desc\}}. I can only perturb exactly \textbf{\{batch len\}} genes at a time. For each predicted perturbation, I am able to measure out the \textbf{\{meas desc\}} which will be referred to as the score. I can only do 5 rounds of experimentation. After every round of experiment, I will provide you with feedback on your predictions, including the correctly identified genes called hits and the corresponding score. The predictions which are not hits will be included in other results.}\end{minipage}}\\
\fbox{\begin{minipage}{\linewidth}
\texttt{
\textbf{USER PROMPT:}\\
This is round \textbf{\{round num\}}.\\
Here is the feedback on all your predictions till now:\\
\textbf{\{feedback\}}\\
Here is a strategy to follow: Update your priors appropriately and choose genes that gave you hits. Also, be sure to explore by including some genes that could give hits.\\
Please propose \textbf{\{num cluster centers\}} different yet valid gene names as per the HGNC nomenclature you want to explore next. Note that I will choose unexplored genes closest to your predicted genes to form the predictions. Your response should exactly follow the format:\\
**Reflection: Thoughts on previous results and next steps.\\
**Research Plan: The full high-level research plan, with current status and reasoning behind each proposed approach. It should be at most 5 sentences.\\
**Solution:\\
\#\# <Gene 1>\\
\#\# <Gene 2>\\
...\\
\#\# <Gene \textbf{\{num cluster centers\}}>\\
Each gene in the solution should only be the gene name in the HGNC nomenclature. DO NOT ADD ANY COMMENTS IN THE SOLUTION OR AFTER THE SOLUTION.
}\end{minipage}}\\

This is the prompt template at any round $i$. If $i =1$, the segment about feedback in the user prompt won't be there. The different fields in bold represent the following:
\begin{itemize}
    \item \textbf{round num}: This is the round number for which experiments are being selected.
    \item \textbf{batch len}: This corresponds to $B$ i.e. the experimental budget in each round.
    \item \textbf{num cluster centers}: This reflects the number of cluster centers that LLMNN generates to sample around.
    \item \textbf{feedback}: This is a string divided into two parts: hits and non-hits. Both parts contain the respective candidate names and their corresponding measurement values.
    \item \textbf{func desc}: This is the description of the task for which we need to perform experiment design. In the context of genes, this means the description of the phenotype we desire to achieve.
    \item \textbf{score desc}: This elaborates on the measurement values we have for the candidates. 
\end{itemize}
Please refer to table \ref{tab:desc-for-genes} for detailed func desc and score desc for each dataset.

\subsection{Chemical Property Optimization}
This prompt is used at any round $i$ for the molecular datasets. Note that for $i=1$, the feedback segment won't be included in the prompt. Most fields described in this prompt are similar to the genetic perturbation, with the exception of \textbf{candidate space info} that contains a high-level description of the candidate space to provide more context to the LLM. Table \ref{tab:desc-for-mols} contains the details of func desc and candidate space info for each of the 3 datasets.
\fbox{\begin{minipage}{\linewidth}
\texttt{
\textbf{SYSTEM PROMPT:}\\
You are a chemistry expert who will assist me with problems in molecular property optimization. Given a library of molecules, I am planning to conduct wet-lab experiments to identify molecules that have high \textbf{\{func desc\}}. \textbf{\{candidate space info\}} I can only experiment with exactly \textbf{\{batch len\}} molecules at a time. For each predicted molecule, I am able to measure out the property value, which will be referred to as the score. I can only do 5 rounds of experimentation. After every round of experiment, I will provide you with feedback on your predictions, including the correctly identified molecules called hits and the corresponding score. The predictions which are not hits will be included in other results.}\end{minipage}}
\fbox{\begin{minipage}{\linewidth}
\texttt{
\textbf{USER PROMPT:}\\
This is round \textbf{\{round num\}}.\\
Here is the feedback on all your predictions till now:
\textbf{\{feedback str\}}\\
Here is a strategy to follow: Update your priors appropriately and choose SMILES that gave you hits. Also, be sure to explore by including some SMILES strings that could give hits.\\
Please propose \textbf{\{num cluster centers\}} different yet valid SMILES strings of molecules you want to explore next. Note that I will choose unexplored molecules closest to your predicted SMILES strings to form the predictions. Your response should exactly follow the format:\\
**Reflection: Thoughts on previous results and next steps.\\
**Research Plan: The full high level research plan, with current status and reasoning behind each proposed approach. It should be at most 5 sentences.\\
**Solution:\\
\#\# <SMILES 1>\\
\#\# <SMILES 2>\\
...\\
\#\# <SMILES \textbf{\{num cluster centers\}}>\\
Each SMILES string in the solution should be a SMILES string representation of a valid molecule.\\
DO NOT ADD ANY COMMENTS IN THE SOLUTION OR AFTER THE SOLUTION.}\end{minipage}}

\section{Experiment Setup Details}
The following subsections describe the additional experiment details for better understanding:

\subsection{Datasets}
\subsubsection{Single Gene Perturbation}
\label{sec:gene-perturb-datasets}
The 5 datasets used in this domain are borrowed from BioDiscoveryAgent \citep{roohani2024biodiscoveryagentaiagentdesigning} and are as follows: 1) \textbf{IL2} \citep{schmidt2022crispr} is a dataset that measures the changes in the production of Interleukin-2 (IL2) cytokine involved in immune signaling; 2) \textbf{IFNG} \citep{schmidt2022crispr} is another similar dataset that measures changes in production of Interferon-$\gamma$ cytokine; 3) \textbf{Carnevale} \citep{carnevale2022rasa2} aims to identify genes that render T cells resistant to inhibitory signals encountered in the tumor microenvironment; 4) \textbf{Sanchez} \citep{sanchez2021genome} dataset studies the change in expression of endogenous tau protein levels in neurons and 5) \textbf{Sanchez Down} \citep{sanchez2021genome} is exactly same as Sanchez dataset but focuses purely on decreasing the expression unlike the Sanchez which includes both increasing and decreasing expression.

\subsubsection{Chemical Property Optimization}
\label{sec:chem-datasets-app}
We describe the molecular property datasets in greater detail below:
\begin{itemize}[leftmargin=*]
    \item \textbf{ESOL:} Part of the MoleculeNet benchmark, ESOL \citep{delaney2004esol} is a small dataset comprising the water solubility data for $\mid \mathcal{C} \mid = 1128$ compounds in log moles per litre. Each compound is represented as a SMILES string.
    \item \textbf{FreeSolv:} Free Solvation (FreeSolv) Database \citep{mobley2014freesolv} is another small dataset that provides experimental and calculated hydration free energies of $\mid \mathcal{C} \mid = 642$ small molecules in water.
    \item \textbf{Ion. E.:} This dataset is a subset of molecular data from the JCESR molecules that are a part of the Materials Project Database \citep{10.1063/1.4812323}. The original set contains approximately $25,000$ molecules with annotations of many attributes and properties. Our experiments focus on one property: the Ionization Energy. We apply two levels of filters on the whole set of $25,000$ molecules: first, we restrict to only those molecules whose ionization energy lies between $-10$ and $10$ units (to avoid noisy outliers), and out of all these filtered molecules, we choose the molecules composed solely of `C', `H', `N' or `O' atoms, leaving us with $\mid \mathcal{C} \mid = 11,565$ candidates. These filters avoid the dominating effects due to particular elements, leaving primary dependence on the molecular structure and functional groups captured in the SMILES strings.
\end{itemize}

\subsection{Compute}
All the experiments in this study have been conducted on a single Nvidia A100 GPU with 40 GB of memory. Since our proposed method, LLMNN, involves inference through the LLMs and repeated retrieval from the candidate memory, on average, a single run takes between 5-10 minutes to run.

\subsection{Embeddings}
\label{sec:embeds-details}
\textbf{Achilles Embeddings.} We use the publicly available Achilles embeddings \citep{tsherniak2017defining} for the gene similarity tool. These embeddings are 808 dimensional in size.

\textbf{Molformer Embeddings.} We use the publicly available Molformer-XL model to embed the SMILES strings\footnote{Model checkpoint taken from \href{https://huggingface.co/ibm-research/MoLFormer-XL-both-10pct}{https://huggingface.co/ibm-research/MoLFormer-XL-both-10pct}}. This model has been trained on a collection of SMILES strings from the ZINC and PubChem datasets to learn molecular representations. These embeddings are 768-dimensional in size.

\textbf{Llama 3.1 Embeddings.} We use the LLM2Vec \citep{behnamghader2024llm2vec} approach to obtain the Llama 3.1 embeddings. Specifically, we use the publicly released checkpoints by the authors on Huggingface \footnote{Model checkpoint taken from \href{https://huggingface.co/McGill-NLP/LLM2Vec-Meta-Llama-31-8B-Instruct-mntp-supervised}{https://huggingface.co/McGill-NLP/LLM2Vec-Meta-Llama-31-8B-Instruct-mntp-supervised}}. These embeddings are 4096-dimensional in nature.

\textbf{Qwen2 Embeddings.} We use the publicly available Qwen2-based general text embedding model released by Alibaba-NLP \citep{li2023towards}, as they claim to be on top of the MTEB leaderboard\footnote{Model checkpoint taken from \href{https://huggingface.co/Alibaba-NLP/gte-Qwen2-7B-instruct}{https://huggingface.co/Alibaba-NLP/gte-Qwen2-7B-instruct}}. These embeddings are 3584 dimensional in nature.

\section{Full trace of LLMNN}
\label{sec:full-trace}
Below we present a full trace of the LLMNN method, including both the prompts and the LLM outputs, on the IL2 dataset in the single gene perturbation domain.

\begin{lstlisting}[caption={Full trace of LLMNN method with explanations on IL2 dataset},label={lst:trace-run},frame=single,breaklines=true,breakindent=0pt,basicstyle=\small\ttfamily]
    ----ROUND 1: BEGIN PROMPT----
  SystemMessage(You are a biomedicine expert who will assist me on problems in drug discovery. I am planning to run a CRISPR screen to identify genes that regulate the production of Interleukin-2 (IL-2). I can only perturb exactly 128 genes at a time. For each predicted perturbation, I am able to measure out the log fold change in Interleukin-2 (IL-2) normalized read counts which will be referred to as the score. I can only do 5 rounds of experimentation. After every round of experiment, I will provide you with feedback on your predictions, including the correctly identified genes called hits and the corresponding score. The predictions which are not hits will be included in other results.)
  HumanMessage(
    This is round 1. We are beginning with our experiments.
    Here is a strategy to follow: Choose genes that are very different in their biological pathways to discover what pathways give you hits.
    Please propose 5 different yet valid gene names as per the HGNC nomenclature you want to explore next. Note that I will choose unexplored genes closest to your predicted genes to form the predictions. Your response should exactly follow the format:
    **Reflection: Thoughts on previous results and next steps.
    **Research Plan: The full high level research plan, with current status and reasoning behind each proposed approach. It should be at most 5 sentences.
    **Solution:
    ## <Gene 1>
    ## <Gene 2>
    ...
    ## <Gene 5>
    Each gene in the solution should only be the gene name in the HGNC nomenclature.
    DO NOT ADD ANY COMMENTS IN THE SOLUTION OR AFTER THE SOLUTION.
  )
    ----END PROMPT----

    ----BEGIN OUTPUT----
    **Reflection: Thoughts on previous results and next steps.**
    Since we are starting from scratch, there is no previous data to reflect on. However, we will follow the strategy of choosing genes that are very different in their biological pathways to increase the chances of discovering novel regulators of IL-2 production.
    
    **Research Plan: The full high level research plan, with current status and reasoning behind each proposed approach.**
    We will choose genes involved in different biological processes such as cell signaling, transcription regulation, and metabolic pathways to identify potential regulators of IL-2 production. For round 1, we will select genes that are involved in the regulation of cell growth, differentiation, and metabolism. We will use the HGNC database to identify genes with diverse functions. Our goal is to identify at least one hit in this round. We will use the provided feedback to refine our strategy for the next round.
    
    **Solution:**
    ## ABL1
    ## HNF4A
    ## MAPK14
    ## PAK4
    ## SMAD2
    ----END OUTPUT----

    ----ROUND 2: BEGIN PROMPT---
  SystemMessage(You are a biomedicine expert who will assist me on problems in drug discovery. I am planning to run a CRISPR screen to identify genes that regulate the production of Interleukin-2 (IL-2). I can only perturb exactly 128 genes at a time. For each predicted perturbation, I am able to measure out the log fold change in Interleukin-2 (IL-2) normalized read counts which will be referred to as the score. I can only do 5 rounds of experimentation. After every round of experiment, I will provide you with feedback on your predictions, including the correctly identified genes called hits and the corresponding score. The predictions which are not hits will be included in other results.)
  HumanMessage(
    This is round 2.
    Here is the feedback on all your predictions till now:
    [HITS]
       name  score
      MYBL2   0.51
       GBF1   0.43
      DDX41   0.47
      ZMAT2   0.45
       RPL4   0.35
      HNF4A  -0.34
      CDC16   0.48
       WDR5   0.82
        MYC   0.41
     ZNF536  -0.38
       TLN1  -0.49
      LIN37  -0.39
    SULT2B1  -0.44
    [OTHER RESULTS]
        name  score
        ABL1   0.09
        QRFP   0.00
        COA3  -0.05
        E2F6  -0.11
       NANOG  -0.05
      NDUFA4   0.04
         RAN   0.12
      MRPL19  -0.14
        SLU7   0.13
        CD8B  -0.01
       HMGB1  -0.12
       AP2S1   0.03
       RPS28  -0.16
       COX5B   0.07
       COPS6  -0.08
       RNGTT   0.03
       RPL11   0.05
       NCAPG   0.17
         NSF  -0.05
       NCBP2   0.00
      OR51L1  -0.17
         GHR   0.02
       TSHZ2  -0.01
        GPX3   0.11
        TOX2   0.10
     FAM107B   0.05
       RGPD3  -0.24
    TRAPPC3L   0.10
       WFDC6  -0.00
     SLC24A3   0.10
        IFNE  -0.04
     ZFAND2A   0.12
      BNIP3L  -0.17
     FAM184B  -0.10
        RGMB   0.02
      ZNF853  -0.09
       NPAS2   0.26
       ATAD1  -0.05
        SMCP  -0.08
      RAB27B   0.14
        BIN3  -0.10
       NPY2R   0.02
        EQTN   0.16
      AKR1B1  -0.21
      MAPK14  -0.02
       MLST8  -0.15
     PPP2R3C  -0.19
       MNAT1   0.07
       NOL10   0.05
        EMC7   0.08
        EMC3  -0.05
        GUK1   0.19
        MED6  -0.16
       CIAO1   0.10
      SNAPC5   0.01
       ZCRB1  -0.03
     TRMT112   0.07
       RPS29   0.23
       PDCD2   0.10
       RPP21  -0.16
        RPL5   0.10
        EIF6   0.19
        SNW1   0.28
      HEATR1   0.06
       RPL28   0.34
       WDR18   0.07
        PAK4   0.07
        AFDN  -0.06
         VCL  -0.02
    CATSPERG  -0.09
        PTK2  -0.10
       BCAR1  -0.08
       ACTR3  -0.06
     RAPGEF1   0.19
       WWTR1  -0.10
        VWA8   0.26
      FERMT2  -0.12
        RSU1   0.12
       MSTO1   0.02
      NLRP2B  -0.09
       ARPC4   0.05
       ARPC2  -0.12
       KIF2C  -0.13
         SRF   0.01
      RNASE1   0.16
      NCKAP1  -0.14
      ZFAND5  -0.01
       SMAD2   0.15
      MAGED2  -0.03
       POTEE  -0.09
      IFNA10  -0.18
      AKAIN1  -0.18
      ZNF614   0.08
      ZNF157  -0.11
       MAPK8   0.09
        POLI  -0.21
      ATXN3L  -0.11
     FOXD4L4  -0.09
        CDR1  -0.12
       NELL1   0.03
       AP1S2   0.06
       IFNB1  -0.13
      ARMCX4  -0.14
      ZNF729  -0.10
      NKX3-2  -0.13
       LUZP4  -0.17
       STX17   0.07
      SPINK2   0.29
       MEIG1   0.07
      DAB2IP  -0.22
      ZNF611   0.03
       VGLL1  -0.05
      OR52E6   0.06
      GUCY2F   0.03
      OR10Z1   0.07
    Here is a strategy to follow: Update your priors appropriately and choose genes that gave you hits. Also, be sure to explore by including some genes that could give hits.
    Please propose 5 different yet valid gene names as per the HGNC nomenclature you want to explore next. Note that I will choose unexplored genes closest to your predicted genes to form the predictions. Your response should exactly follow the format:
    **Reflection: Thoughts on previous results and next steps.
    **Research Plan: The full high level research plan, with current status and reasoning behind each proposed approach. It should be at most 5 sentences.
    **Solution:
    ## <Gene 1>
    ## <Gene 2>
    ...
    ## <Gene 5>
    Each gene in the solution should only be the gene name in the HGNC nomenclature.
    DO NOT ADD ANY COMMENTS IN THE SOLUTION OR AFTER THE SOLUTION.
  )
    ----END PROMPT----
    ----BEGIN OUTPUT----
    **Reflection: Thoughts on previous results and next steps.**

    The previous round of CRISPR screen identified 11 hits that regulate the production of Interleukin-2 (IL-2), including MYBL2, GBF1, DDX41, ZMAT2, RPL4, HNF4A, CDC16, WDR5, MYC, ZNF536, and TLN1. The scores of these hits ranged from 0.35 to 0.82, indicating varying degrees of regulation. The other results included 64 genes that did not show significant regulation of IL-2 production. To refine our understanding of the IL-2 regulatory network, we will focus on the hits and their neighbors. We will update our priors to reflect the observed hits and select new genes that are closely related to the identified hits.
    
    **Research Plan: The full high-level research plan, with current status and reasoning behind each proposed approach.**
    
    We will update our prior probabilities based on the hits identified in the previous round and select new genes that are closely related to the hits. We will choose genes that are in the same biological pathways, have similar functional annotations, or have known interactions with the hits. We will then select the top-scoring genes from these candidates to form our next set of predictions. Our goal is to refine our understanding of the IL-2 regulatory network and identify new genes that play a role in its regulation. We will repeat this process for a total of 5 rounds, refining our predictions and updating our priors after each round.
    
    **Solution:**
    ## MYBL2
    ## GBF1
    ## DDX41
    ## ZMAT2
    ## RPL4
    ----END OUTPUT----
    ----ROUND 3: BEGIN PROMPT----
  SystemMessage(You are a biomedicine expert who will assist me on problems in drug discovery. I am planning to run a CRISPR screen to identify genes that regulate the production of Interleukin-2 (IL-2). I can only perturb exactly 128 genes at a time. For each predicted perturbation, I am able to measure out the log fold change in Interleukin-2 (IL-2) normalized read counts which will be referred to as the score. I can only do 5 rounds of experimentation. After every round of experiment, I will provide you with feedback on your predictions, including the correctly identified genes called hits and the corresponding score. The predictions which are not hits will be included in other results.)
  HumanMessage(
    This is round 3.
    Here is the feedback on all your predictions till now:
    [HITS]
        name  score
       MYBL2   0.51
        GBF1   0.43
       DDX41   0.47
       ZMAT2   0.45
        RPL4   0.35
       HNF4A  -0.34
       CDC16   0.48
        WDR5   0.82
         MYC   0.41
      ZNF536  -0.38
        TLN1  -0.49
       LIN37  -0.39
     SULT2B1  -0.44
       SPC24   0.58
      SNRPD2   0.39
      POLR1A   0.47
       SF3A2   0.65
    SNRNP200   0.78
        CDK1   0.43
         DUT   0.44
        PRC1   0.38
      DONSON   0.65
        EEF2   0.43
       PSMA6   0.34
        XAB2   0.41
         VCP   0.66
       PSMA1   0.39
       PSMD7   0.57
        SMU1   0.45
        POLE   0.57
       SF3B3   0.36
       KIF11   0.36
        ETF1  -0.45
       CDC23   0.75
       GINS2   0.38
       RPS11   0.64
       RPS20   0.43
       RBM22   0.61
         FAU   0.49
       RPL17   0.49
      PRPF19   0.38
        RPL9   0.44
       UTP15   0.59
       RPS3A   0.39
       RPL37   0.68
      RPS27A   0.48
       RPL27   0.87
       PRPF8   0.45
        CCT4   0.35
       RPS16   0.38
       RPL7A   0.61
        CDC7   0.62
        RPS6   0.47
      RPL23A   0.79
       RPS13   0.92
       RPL32   0.68
       RPL18   0.77
       RPLP2   0.52
        RPS8   0.69
      RPL10A   0.86
        RPL8   0.81
       MAK16   1.15
       RPS17   0.82
       RPL19   1.21
        RPL3   0.42
       RPL15   0.92
       CPSF4   1.13
        XPO1   0.35
        RRM2   0.50
       RPL14   0.61
    [OTHER RESULTS]
        name  score
        ABL1   0.09
        QRFP   0.00
        COA3  -0.05
        E2F6  -0.11
       NANOG  -0.05
      NDUFA4   0.04
         RAN   0.12
      MRPL19  -0.14
        SLU7   0.13
        CD8B  -0.01
       HMGB1  -0.12
       AP2S1   0.03
       RPS28  -0.16
       COX5B   0.07
       COPS6  -0.08
       RNGTT   0.03
       RPL11   0.05
       NCAPG   0.17
         NSF  -0.05
       NCBP2   0.00
      OR51L1  -0.17
         GHR   0.02
       TSHZ2  -0.01
        GPX3   0.11
        TOX2   0.10
     FAM107B   0.05
       RGPD3  -0.24
    TRAPPC3L   0.10
       WFDC6  -0.00
     SLC24A3   0.10
        IFNE  -0.04
     ZFAND2A   0.12
      BNIP3L  -0.17
     FAM184B  -0.10
        RGMB   0.02
      ZNF853  -0.09
       NPAS2   0.26
       ATAD1  -0.05
        SMCP  -0.08
      RAB27B   0.14
        BIN3  -0.10
       NPY2R   0.02
        EQTN   0.16
      AKR1B1  -0.21
      MAPK14  -0.02
       MLST8  -0.15
     PPP2R3C  -0.19
       MNAT1   0.07
       NOL10   0.05
        EMC7   0.08
        EMC3  -0.05
        GUK1   0.19
        MED6  -0.16
       CIAO1   0.10
      SNAPC5   0.01
       ZCRB1  -0.03
     TRMT112   0.07
       RPS29   0.23
       PDCD2   0.10
       RPP21  -0.16
        RPL5   0.10
        EIF6   0.19
        SNW1   0.28
      HEATR1   0.06
       RPL28   0.34
       WDR18   0.07
        PAK4   0.07
        AFDN  -0.06
         VCL  -0.02
    CATSPERG  -0.09
        PTK2  -0.10
       BCAR1  -0.08
       ACTR3  -0.06
     RAPGEF1   0.19
       WWTR1  -0.10
        VWA8   0.26
      FERMT2  -0.12
        RSU1   0.12
       MSTO1   0.02
      NLRP2B  -0.09
       ARPC4   0.05
       ARPC2  -0.12
       KIF2C  -0.13
         SRF   0.01
      RNASE1   0.16
      NCKAP1  -0.14
      ZFAND5  -0.01
       SMAD2   0.15
      MAGED2  -0.03
       POTEE  -0.09
      IFNA10  -0.18
      AKAIN1  -0.18
      ZNF614   0.08
      ZNF157  -0.11
       MAPK8   0.09
        POLI  -0.21
      ATXN3L  -0.11
     FOXD4L4  -0.09
        CDR1  -0.12
       NELL1   0.03
       AP1S2   0.06
       IFNB1  -0.13
      ARMCX4  -0.14
      ZNF729  -0.10
      NKX3-2  -0.13
       LUZP4  -0.17
       STX17   0.07
      SPINK2   0.29
       MEIG1   0.07
      DAB2IP  -0.22
      ZNF611   0.03
       VGLL1  -0.05
      OR52E6   0.06
      GUCY2F   0.03
      OR10Z1   0.07
        CLTC   0.26
       THOC7   0.17
      ANKLE2   0.20
       SF3A1   0.17
     SAP30BP   0.26
       ZMAT5   0.19
        LSM2  -0.18
       CDC45   0.08
      TANGO6   0.10
       NUP85  -0.11
     TBC1D3B   0.13
       SEC13   0.19
       NEDD1   0.32
        RFC2   0.30
        SDE2   0.31
        CCT3   0.23
       PSMB3  -0.01
       KPNB1   0.14
     ANAPC11  -0.07
       HSPA9  -0.03
        SMG1   0.15
       SBNO1  -0.03
       PSMB4   0.14
       COPB1   0.14
       PSMA3   0.05
       COPB2   0.24
       PSMA5   0.34
       PSMA2   0.29
      CDK11A  -0.06
     PRELID1   0.22
       FARSB   0.13
     RSL24D1   0.11
       RPL12   0.34
       DHX37   0.29
       BUD31   0.08
      CDC123  -0.03
      TXNL4A  -0.25
      INTS11   0.26
        DBR1   0.14
       HSPE1  -0.03
      NPIPB6  -0.15
    PAFAH1B1  -0.26
        NFS1   0.15
       DDX56   0.21
       CDC27   0.17
       DDX49   0.22
       SF3B4   0.18
       RPL23   0.09
        WEE1  -0.18
        NIP7   0.22
       SF3B5   0.26
       CDC20   0.29
        RPS9   0.12
        LRR1  -0.03
      GEMIN5   0.00
      SNRPA1   0.06
       RPLP0   0.26
      EIF2S1   0.02
     DNAJC17   0.04
       XRCC6  -0.22
        DUX4  -0.01
       RPS19   0.23
       WDR43   0.24
        RRM1   0.32
     ATP6V0C   0.06
        RPS2   0.33
       PUF60  -0.07
     SEC61A1  -0.17
        TSR1   0.23
       WDR74   0.33
        RPL7   0.31
    Here is a strategy to follow: Update your priors appropriately and choose genes that gave you hits. Also, be sure to explore by including some genes that could give hits.
    Please propose 5 different yet valid gene names as per the HGNC nomenclature you want to explore next. Note that I will choose unexplored genes closest to your predicted genes to form the predictions. Your response should exactly follow the format:
    **Reflection: Thoughts on previous results and next steps.
    **Research Plan: The full high level research plan, with current status and reasoning behind each proposed approach. It should be at most 5 sentences.
    **Solution:
    ## <Gene 1>
    ## <Gene 2>
    ...
    ## <Gene 5>
    Each gene in the solution should only be the gene name in the HGNC nomenclature.
    DO NOT ADD ANY COMMENTS IN THE SOLUTION OR AFTER THE SOLUTION.
  )
    ----END PROMPT----
    ----BEGIN OUTPUT----
    **Reflection: Thoughts on previous results and next steps.**
    Based on the previous round of experimentation, we have identified several genes that are associated with the regulation of Interleukin-2 (IL-2) production. The hits from this round include genes involved in various cellular processes such as RNA processing, translation, and cell cycle regulation. We can update our priors to reflect the importance of these genes in IL-2 regulation. To further explore the regulatory network, we will choose genes that are functionally related to the hits from the previous round.
    
    **Research Plan: The full high-level research plan, with current status and reasoning behind each proposed approach.**
    We will select genes that are functionally related to the hits from the previous round, focusing on RNA processing, translation, and cell cycle regulation. We will choose unexplored genes closest to the hits to form the predictions. This approach will allow us to further refine our understanding of the IL-2 regulatory network. We will use the updated priors to inform our selection of genes. Our goal is to identify additional genes that regulate IL-2 production.
    
    **Solution:**
    ## RPS27
    ## SF3B1
    ## DDX3X
    ## RPS15
    ## NOLC1
    ----END OUTPUT----
    ----ROUND 4: BEGIN PROMPT----
  SystemMessage(You are a biomedicine expert who will assist me on problems in drug discovery. I am planning to run a CRISPR screen to identify genes that regulate the production of Interleukin-2 (IL-2). I can only perturb exactly 128 genes at a time. For each predicted perturbation, I am able to measure out the log fold change in Interleukin-2 (IL-2) normalized read counts which will be referred to as the score. I can only do 5 rounds of experimentation. After every round of experiment, I will provide you with feedback on your predictions, including the correctly identified genes called hits and the corresponding score. The predictions which are not hits will be included in other results.)
  HumanMessage(
    This is round 4.
    Here is the feedback on all your predictions till now:
    [HITS]
        name  score
       MYBL2   0.51
        GBF1   0.43
       DDX41   0.47
       ZMAT2   0.45
        RPL4   0.35
       HNF4A  -0.34
       CDC16   0.48
        WDR5   0.82
         MYC   0.41
      ZNF536  -0.38
        TLN1  -0.49
       LIN37  -0.39
     SULT2B1  -0.44
       SPC24   0.58
      SNRPD2   0.39
      POLR1A   0.47
       SF3A2   0.65
    SNRNP200   0.78
        CDK1   0.43
         DUT   0.44
        PRC1   0.38
      DONSON   0.65
        EEF2   0.43
       PSMA6   0.34
        XAB2   0.41
         VCP   0.66
       PSMA1   0.39
       PSMD7   0.57
        SMU1   0.45
        POLE   0.57
       SF3B3   0.36
       KIF11   0.36
        ETF1  -0.45
       CDC23   0.75
       GINS2   0.38
       RPS11   0.64
       RPS20   0.43
       RBM22   0.61
         FAU   0.49
       RPL17   0.49
      PRPF19   0.38
        RPL9   0.44
       UTP15   0.59
       RPS3A   0.39
       RPL37   0.68
      RPS27A   0.48
       RPL27   0.87
       PRPF8   0.45
        CCT4   0.35
       RPS16   0.38
       RPL7A   0.61
        CDC7   0.62
        RPS6   0.47
      RPL23A   0.79
       RPS13   0.92
       RPL32   0.68
       RPL18   0.77
       RPLP2   0.52
        RPS8   0.69
      RPL10A   0.86
        RPL8   0.81
       MAK16   1.15
       RPS17   0.82
       RPL19   1.21
        RPL3   0.42
       RPL15   0.92
       CPSF4   1.13
        XPO1   0.35
        RRM2   0.50
       RPL14   0.61
       INTS3   0.53
        RFC3   0.50
       RBM25   0.47
        FCF1   0.40
      POLR1B   0.36
       POLD3   0.52
       CHMP6   0.85
       PSMC5   0.34
       RPL31   0.40
         ERH   0.45
       RPS4X   0.70
       CHERP   0.49
        DKC1   0.40
       CNOT3   0.44
       SNRPC   1.25
        MTBP   0.57
       SYMPK   0.68
        CDC6   0.47
        PPAN   0.89
      SPOUT1   0.38
       EIF3I   0.45
       RPL36   0.72
       NUP93   0.35
       RPS24   0.83
      NUP133   0.49
       RPS18   0.46
       RPS14   0.60
      PDCD11   0.43
       NOC3L   0.34
        BMS1   0.56
       RPS25   0.49
    EBNA1BP2   0.39
        BOP1   0.51
        NOP2   0.55
        RRN3   0.35
     TUBGCP2   0.59
        RPS3   0.54
    [OTHER RESULTS]
        name  score
        ABL1   0.09
        QRFP   0.00
        COA3  -0.05
        E2F6  -0.11
       NANOG  -0.05
      NDUFA4   0.04
         RAN   0.12
      MRPL19  -0.14
        SLU7   0.13
        CD8B  -0.01
       HMGB1  -0.12
       AP2S1   0.03
       RPS28  -0.16
       COX5B   0.07
       COPS6  -0.08
       RNGTT   0.03
       RPL11   0.05
       NCAPG   0.17
         NSF  -0.05
       NCBP2   0.00
      OR51L1  -0.17
         GHR   0.02
       TSHZ2  -0.01
        GPX3   0.11
        TOX2   0.10
     FAM107B   0.05
       RGPD3  -0.24
    TRAPPC3L   0.10
       WFDC6  -0.00
     SLC24A3   0.10
        IFNE  -0.04
     ZFAND2A   0.12
      BNIP3L  -0.17
     FAM184B  -0.10
        RGMB   0.02
      ZNF853  -0.09
       NPAS2   0.26
       ATAD1  -0.05
        SMCP  -0.08
      RAB27B   0.14
        BIN3  -0.10
       NPY2R   0.02
        EQTN   0.16
      AKR1B1  -0.21
      MAPK14  -0.02
       MLST8  -0.15
     PPP2R3C  -0.19
       MNAT1   0.07
       NOL10   0.05
        EMC7   0.08
        EMC3  -0.05
        GUK1   0.19
        MED6  -0.16
       CIAO1   0.10
      SNAPC5   0.01
       ZCRB1  -0.03
     TRMT112   0.07
       RPS29   0.23
       PDCD2   0.10
       RPP21  -0.16
        RPL5   0.10
        EIF6   0.19
        SNW1   0.28
      HEATR1   0.06
       RPL28   0.34
       WDR18   0.07
        PAK4   0.07
        AFDN  -0.06
         VCL  -0.02
    CATSPERG  -0.09
        PTK2  -0.10
       BCAR1  -0.08
       ACTR3  -0.06
     RAPGEF1   0.19
       WWTR1  -0.10
        VWA8   0.26
      FERMT2  -0.12
        RSU1   0.12
       MSTO1   0.02
      NLRP2B  -0.09
       ARPC4   0.05
       ARPC2  -0.12
       KIF2C  -0.13
         SRF   0.01
      RNASE1   0.16
      NCKAP1  -0.14
      ZFAND5  -0.01
       SMAD2   0.15
      MAGED2  -0.03
       POTEE  -0.09
      IFNA10  -0.18
      AKAIN1  -0.18
      ZNF614   0.08
      ZNF157  -0.11
       MAPK8   0.09
        POLI  -0.21
      ATXN3L  -0.11
     FOXD4L4  -0.09
        CDR1  -0.12
       NELL1   0.03
       AP1S2   0.06
       IFNB1  -0.13
      ARMCX4  -0.14
      ZNF729  -0.10
      NKX3-2  -0.13
       LUZP4  -0.17
       STX17   0.07
      SPINK2   0.29
       MEIG1   0.07
      DAB2IP  -0.22
      ZNF611   0.03
       VGLL1  -0.05
      OR52E6   0.06
      GUCY2F   0.03
      OR10Z1   0.07
        CLTC   0.26
       THOC7   0.17
      ANKLE2   0.20
       SF3A1   0.17
     SAP30BP   0.26
       ZMAT5   0.19
        LSM2  -0.18
       CDC45   0.08
      TANGO6   0.10
       NUP85  -0.11
     TBC1D3B   0.13
       SEC13   0.19
       NEDD1   0.32
        RFC2   0.30
        SDE2   0.31
        CCT3   0.23
       PSMB3  -0.01
       KPNB1   0.14
     ANAPC11  -0.07
       HSPA9  -0.03
        SMG1   0.15
       SBNO1  -0.03
       PSMB4   0.14
       COPB1   0.14
       PSMA3   0.05
       COPB2   0.24
       PSMA5   0.34
       PSMA2   0.29
      CDK11A  -0.06
     PRELID1   0.22
       FARSB   0.13
     RSL24D1   0.11
       RPL12   0.34
       DHX37   0.29
       BUD31   0.08
      CDC123  -0.03
      TXNL4A  -0.25
      INTS11   0.26
        DBR1   0.14
       HSPE1  -0.03
      NPIPB6  -0.15
    PAFAH1B1  -0.26
        NFS1   0.15
       DDX56   0.21
       CDC27   0.17
       DDX49   0.22
       SF3B4   0.18
       RPL23   0.09
        WEE1  -0.18
        NIP7   0.22
       SF3B5   0.26
       CDC20   0.29
        RPS9   0.12
        LRR1  -0.03
      GEMIN5   0.00
      SNRPA1   0.06
       RPLP0   0.26
      EIF2S1   0.02
     DNAJC17   0.04
       XRCC6  -0.22
        DUX4  -0.01
       RPS19   0.23
       WDR43   0.24
        RRM1   0.32
     ATP6V0C   0.06
        RPS2   0.33
       PUF60  -0.07
     SEC61A1  -0.17
        TSR1   0.23
       WDR74   0.33
        RPL7   0.31
       RPS27   0.13
        MCM7   0.01
      LUC7L3   0.12
      EEF1A1   0.21
        RPSA   0.28
       PPWD1   0.29
      TOMM22  -0.05
     RACGAP1   0.25
     DYNC1H1  -0.17
     SNRNP25   0.01
    TIMELESS   0.11
       UQCRH  -0.09
       PAM16   0.19
       PFDN6  -0.21
       DDX10   0.32
       RRP12   0.16
        GPN1   0.08
       PFDN2   0.06
      ZNHIT2   0.20
        BYSL   0.04
       USP36  -0.03
       SF3B1   0.30
       VPS25  -0.08
     PRPF38A   0.15
        EIF5  -0.12
     ATP6V0B   0.06
       RNPC3  -0.02
       PSMA4   0.23
        UBL5  -0.00
      EIF4A3   0.33
       POLE2   0.09
        GPN3   0.15
       PSMD3   0.27
        COPA   0.05
      CHMP2A   0.06
      RPL27A   0.27
       PSMD1   0.32
       CWC22   0.22
       WBP11   0.13
       PSMD4   0.23
       DDX3X   0.03
      EIF1AX   0.10
        UBA1   0.21
      EIF2S3   0.10
        TSR2   0.19
       POLA1   0.24
       DHDDS  -0.26
      ANAPC2   0.23
       GPKOW   0.07
       HCFC1   0.06
       PDRG1  -0.16
     TRAPPC3   0.07
        LSM3   0.05
     ATP6AP2  -0.05
        GPN2   0.03
         DTL   0.06
     ZFP36L2   0.01
      RPS15A   0.21
       NEDD8   0.09
       WDR75   0.31
        NSA2   0.12
       TRNT1   0.09
        ISCU  -0.01
      MRPL13   0.09
       DDX55   0.01
       BIRC5   0.30
        ARL2   0.02
        PWP2   0.02
      SPPL2B   0.01
        NIFK   0.26
       ABCE1   0.28
       PRIM1   0.11
       EIF3G   0.08
       NOLC1   0.02
       NOC4L   0.11
       UTP20   0.33
       NOP10  -0.04
        URB1   0.04
        NMD3   0.33
        ZPR1   0.15
        NOL9   0.11
        TCP1   0.02
        TBCB   0.10
        UBA2  -0.17
      RSL1D1   0.28
     PAK1IP1   0.22
       NOP56  -0.08
       RBM17   0.00
        XRN2   0.06
       OSGEP   0.10
       UTP18   0.02
    Here is a strategy to follow: Update your priors appropriately and choose genes that gave you hits. Also, be sure to explore by including some genes that could give hits.
    Please propose 5 different yet valid gene names as per the HGNC nomenclature you want to explore next. Note that I will choose unexplored genes closest to your predicted genes to form the predictions. Your response should exactly follow the format:
    **Reflection: Thoughts on previous results and next steps.
    **Research Plan: The full high level research plan, with current status and reasoning behind each proposed approach. It should be at most 5 sentences.
    **Solution:
    ## <Gene 1>
    ## <Gene 2>
    ...
    ## <Gene 5>
    Each gene in the solution should only be the gene name in the HGNC nomenclature.
    DO NOT ADD ANY COMMENTS IN THE SOLUTION OR AFTER THE SOLUTION.
  )
    ----END PROMPT----
    ----BEGIN OUTPUT----
    **Reflection: Thoughts on previous results and next steps.**
    The previous results have identified several genes that are involved in the regulation of Interleukin-2 (IL-2) production. The top-scoring genes from the previous round include RPL27, RPL23A, RPL18, RPS13, RPL10A, RPL8, MAK16, RPL19, CPSF4, and RPL15. These genes are mostly ribosomal proteins, suggesting that the regulation of IL-2 production is tightly linked to protein synthesis. The next steps will focus on exploring more genes that are closely related to these hits and are not yet explored.
    
    **Research Plan: The full high level research plan, with current status and reasoning behind each proposed approach.**
    We will update our priors based on the previous results and choose genes that are closely related to the hits from the previous round. We will also include some genes that could give hits based on their functional annotation. We will use the HGNC nomenclature to select 5 unexplored genes that are closest to the predicted genes. We will then use these genes for the next round of experimentation.
    
    **Solution:**
    ## RPL22
    ## RPS11
    ## RPL14
    ## RPS4X
    ## RPL32
    ----END OUTPUT----
    ----ROUND 5: BEGIN PROMPT----
    SystemMessage(You are a biomedicine expert who will assist me on problems in drug discovery. I am planning to run a CRISPR screen to identify genes that regulate the production of Interleukin-2 (IL-2). I can only perturb exactly 128 genes at a time. For each predicted perturbation, I am able to measure out the log fold change in Interleukin-2 (IL-2) normalized read counts which will be referred to as the score. I can only do 5 rounds of experimentation. After every round of experiment, I will provide you with feedback on your predictions, including the correctly identified genes called hits and the corresponding score. The predictions which are not hits will be included in other results.)
  HumanMessage(
    This is round 5.
    Here is the feedback on all your predictions till now:
    [HITS]
         name  score
        MYBL2   0.51
         GBF1   0.43
        DDX41   0.47
        ZMAT2   0.45
         RPL4   0.35
        HNF4A  -0.34
        CDC16   0.48
         WDR5   0.82
          MYC   0.41
       ZNF536  -0.38
         TLN1  -0.49
        LIN37  -0.39
      SULT2B1  -0.44
        SPC24   0.58
       SNRPD2   0.39
       POLR1A   0.47
        SF3A2   0.65
     SNRNP200   0.78
         CDK1   0.43
          DUT   0.44
         PRC1   0.38
       DONSON   0.65
         EEF2   0.43
        PSMA6   0.34
         XAB2   0.41
          VCP   0.66
        PSMA1   0.39
        PSMD7   0.57
         SMU1   0.45
         POLE   0.57
        SF3B3   0.36
        KIF11   0.36
         ETF1  -0.45
        CDC23   0.75
        GINS2   0.38
        RPS11   0.64
        RPS20   0.43
        RBM22   0.61
          FAU   0.49
        RPL17   0.49
       PRPF19   0.38
         RPL9   0.44
        UTP15   0.59
        RPS3A   0.39
        RPL37   0.68
       RPS27A   0.48
        RPL27   0.87
        PRPF8   0.45
         CCT4   0.35
        RPS16   0.38
        RPL7A   0.61
         CDC7   0.62
         RPS6   0.47
       RPL23A   0.79
        RPS13   0.92
        RPL32   0.68
        RPL18   0.77
        RPLP2   0.52
         RPS8   0.69
       RPL10A   0.86
         RPL8   0.81
        MAK16   1.15
        RPS17   0.82
        RPL19   1.21
         RPL3   0.42
        RPL15   0.92
        CPSF4   1.13
         XPO1   0.35
         RRM2   0.50
        RPL14   0.61
        INTS3   0.53
         RFC3   0.50
        RBM25   0.47
         FCF1   0.40
       POLR1B   0.36
        POLD3   0.52
        CHMP6   0.85
        PSMC5   0.34
        RPL31   0.40
          ERH   0.45
        RPS4X   0.70
        CHERP   0.49
         DKC1   0.40
        CNOT3   0.44
        SNRPC   1.25
         MTBP   0.57
        SYMPK   0.68
         CDC6   0.47
         PPAN   0.89
       SPOUT1   0.38
        EIF3I   0.45
        RPL36   0.72
        NUP93   0.35
        RPS24   0.83
       NUP133   0.49
        RPS18   0.46
        RPS14   0.60
       PDCD11   0.43
        NOC3L   0.34
         BMS1   0.56
        RPS25   0.49
     EBNA1BP2   0.39
         BOP1   0.51
         NOP2   0.55
         RRN3   0.35
      TUBGCP2   0.59
         RPS3   0.54
       PKMYT1   0.46
       CLNS1A   0.47
        GINS1   0.47
      SNRNP27   0.63
        TNPO3   0.44
         MCM5   0.43
         BUB3   0.48
        WDR12   0.43
        NUTF2   0.45
        RPLP1   0.66
      DYNC1I2   0.62
         USP5   0.34
        RPL35   0.98
        RPL26   0.99
        GINS4   0.59
        EIF3B   0.48
        GNL3L   0.39
        SMC1A   0.68
        RPL34   0.67
         DAD1  -0.42
    MPHOSPH10   0.47
         RPL6   0.62
         GNL2   0.42
       RUVBL1   0.56
         RPS7   1.02
        SNRPF   0.57
        MFAP1   0.42
        SRBD1   0.56
       POLR1C   0.41
         NOB1   0.38
        PSMD6   0.43
    [OTHER RESULTS]
        name  score
        ABL1   0.09
        QRFP   0.00
        COA3  -0.05
        E2F6  -0.11
       NANOG  -0.05
      NDUFA4   0.04
         RAN   0.12
      MRPL19  -0.14
        SLU7   0.13
        CD8B  -0.01
       HMGB1  -0.12
       AP2S1   0.03
       RPS28  -0.16
       COX5B   0.07
       COPS6  -0.08
       RNGTT   0.03
       RPL11   0.05
       NCAPG   0.17
         NSF  -0.05
       NCBP2   0.00
      OR51L1  -0.17
         GHR   0.02
       TSHZ2  -0.01
        GPX3   0.11
        TOX2   0.10
     FAM107B   0.05
       RGPD3  -0.24
    TRAPPC3L   0.10
       WFDC6  -0.00
     SLC24A3   0.10
        IFNE  -0.04
     ZFAND2A   0.12
      BNIP3L  -0.17
     FAM184B  -0.10
        RGMB   0.02
      ZNF853  -0.09
       NPAS2   0.26
       ATAD1  -0.05
        SMCP  -0.08
      RAB27B   0.14
        BIN3  -0.10
       NPY2R   0.02
        EQTN   0.16
      AKR1B1  -0.21
      MAPK14  -0.02
       MLST8  -0.15
     PPP2R3C  -0.19
       MNAT1   0.07
       NOL10   0.05
        EMC7   0.08
        EMC3  -0.05
        GUK1   0.19
        MED6  -0.16
       CIAO1   0.10
      SNAPC5   0.01
       ZCRB1  -0.03
     TRMT112   0.07
       RPS29   0.23
       PDCD2   0.10
       RPP21  -0.16
        RPL5   0.10
        EIF6   0.19
        SNW1   0.28
      HEATR1   0.06
       RPL28   0.34
       WDR18   0.07
        PAK4   0.07
        AFDN  -0.06
         VCL  -0.02
    CATSPERG  -0.09
        PTK2  -0.10
       BCAR1  -0.08
       ACTR3  -0.06
     RAPGEF1   0.19
       WWTR1  -0.10
        VWA8   0.26
      FERMT2  -0.12
        RSU1   0.12
       MSTO1   0.02
      NLRP2B  -0.09
       ARPC4   0.05
       ARPC2  -0.12
       KIF2C  -0.13
         SRF   0.01
      RNASE1   0.16
      NCKAP1  -0.14
      ZFAND5  -0.01
       SMAD2   0.15
      MAGED2  -0.03
       POTEE  -0.09
      IFNA10  -0.18
      AKAIN1  -0.18
      ZNF614   0.08
      ZNF157  -0.11
       MAPK8   0.09
        POLI  -0.21
      ATXN3L  -0.11
     FOXD4L4  -0.09
        CDR1  -0.12
       NELL1   0.03
       AP1S2   0.06
       IFNB1  -0.13
      ARMCX4  -0.14
      ZNF729  -0.10
      NKX3-2  -0.13
       LUZP4  -0.17
       STX17   0.07
      SPINK2   0.29
       MEIG1   0.07
      DAB2IP  -0.22
      ZNF611   0.03
       VGLL1  -0.05
      OR52E6   0.06
      GUCY2F   0.03
      OR10Z1   0.07
        CLTC   0.26
       THOC7   0.17
      ANKLE2   0.20
       SF3A1   0.17
     SAP30BP   0.26
       ZMAT5   0.19
        LSM2  -0.18
       CDC45   0.08
      TANGO6   0.10
       NUP85  -0.11
     TBC1D3B   0.13
       SEC13   0.19
       NEDD1   0.32
        RFC2   0.30
        SDE2   0.31
        CCT3   0.23
       PSMB3  -0.01
       KPNB1   0.14
     ANAPC11  -0.07
       HSPA9  -0.03
        SMG1   0.15
       SBNO1  -0.03
       PSMB4   0.14
       COPB1   0.14
       PSMA3   0.05
       COPB2   0.24
       PSMA5   0.34
       PSMA2   0.29
      CDK11A  -0.06
     PRELID1   0.22
       FARSB   0.13
     RSL24D1   0.11
       RPL12   0.34
       DHX37   0.29
       BUD31   0.08
      CDC123  -0.03
      TXNL4A  -0.25
      INTS11   0.26
        DBR1   0.14
       HSPE1  -0.03
      NPIPB6  -0.15
    PAFAH1B1  -0.26
        NFS1   0.15
       DDX56   0.21
       CDC27   0.17
       DDX49   0.22
       SF3B4   0.18
       RPL23   0.09
        WEE1  -0.18
        NIP7   0.22
       SF3B5   0.26
       CDC20   0.29
        RPS9   0.12
        LRR1  -0.03
      GEMIN5   0.00
      SNRPA1   0.06
       RPLP0   0.26
      EIF2S1   0.02
     DNAJC17   0.04
       XRCC6  -0.22
        DUX4  -0.01
       RPS19   0.23
       WDR43   0.24
        RRM1   0.32
     ATP6V0C   0.06
        RPS2   0.33
       PUF60  -0.07
     SEC61A1  -0.17
        TSR1   0.23
       WDR74   0.33
        RPL7   0.31
       RPS27   0.13
        MCM7   0.01
      LUC7L3   0.12
      EEF1A1   0.21
        RPSA   0.28
       PPWD1   0.29
      TOMM22  -0.05
     RACGAP1   0.25
     DYNC1H1  -0.17
     SNRNP25   0.01
    TIMELESS   0.11
       UQCRH  -0.09
       PAM16   0.19
       PFDN6  -0.21
       DDX10   0.32
       RRP12   0.16
        GPN1   0.08
       PFDN2   0.06
      ZNHIT2   0.20
        BYSL   0.04
       USP36  -0.03
       SF3B1   0.30
       VPS25  -0.08
     PRPF38A   0.15
        EIF5  -0.12
     ATP6V0B   0.06
       RNPC3  -0.02
       PSMA4   0.23
        UBL5  -0.00
      EIF4A3   0.33
       POLE2   0.09
        GPN3   0.15
       PSMD3   0.27
        COPA   0.05
      CHMP2A   0.06
      RPL27A   0.27
       PSMD1   0.32
       CWC22   0.22
       WBP11   0.13
       PSMD4   0.23
       DDX3X   0.03
      EIF1AX   0.10
        UBA1   0.21
      EIF2S3   0.10
        TSR2   0.19
       POLA1   0.24
       DHDDS  -0.26
      ANAPC2   0.23
       GPKOW   0.07
       HCFC1   0.06
       PDRG1  -0.16
     TRAPPC3   0.07
        LSM3   0.05
     ATP6AP2  -0.05
        GPN2   0.03
         DTL   0.06
     ZFP36L2   0.01
      RPS15A   0.21
       NEDD8   0.09
       WDR75   0.31
        NSA2   0.12
       TRNT1   0.09
        ISCU  -0.01
      MRPL13   0.09
       DDX55   0.01
       BIRC5   0.30
        ARL2   0.02
        PWP2   0.02
      SPPL2B   0.01
        NIFK   0.26
       ABCE1   0.28
       PRIM1   0.11
       EIF3G   0.08
       NOLC1   0.02
       NOC4L   0.11
       UTP20   0.33
       NOP10  -0.04
        URB1   0.04
        NMD3   0.33
        ZPR1   0.15
        NOL9   0.11
        TCP1   0.02
        TBCB   0.10
        UBA2  -0.17
      RSL1D1   0.28
     PAK1IP1   0.22
       NOP56  -0.08
       RBM17   0.00
        XRN2   0.06
       OSGEP   0.10
       UTP18   0.02
       RPL22  -0.23
     TP53BP1  -0.04
      NELFCD   0.03
     METTL14   0.14
      DEPDC1   0.00
        WTAP   0.12
         CA6  -0.08
        PREB   0.13
      ZNF676  -0.04
    PRAMEF18  -0.11
        STN1   0.01
     SLC35G6  -0.20
      ZNF318  -0.13
       DPPA2   0.18
       GSTM3   0.11
      SRSF10   0.13
       SCAF4  -0.04
       PNRC2   0.07
      IFITM3   0.00
       WDR55   0.11
        TMA7  -0.06
       MRPS2   0.14
      TSEN54   0.21
       GGPS1   0.06
      EIF1AD   0.19
      RPL18A   0.31
        SAE1  -0.03
     RABGGTA   0.25
      SS18L2   0.10
      MRPS12   0.06
       HAUS5   0.25
       FOXD4   0.15
       TOP2A   0.13
     SNRNP70   0.33
       INTS9   0.01
       NOP58   0.22
       SSU72   0.13
      TOMM40  -0.06
       HSPD1  -0.02
        YKT6  -0.04
     ATP6V1A  -0.07
        LSM7   0.09
     ATP6V1F   0.03
       CSE1L   0.06
       RBM19   0.14
        RPA3  -0.02
       PSMB2   0.17
        LSM8   0.16
       SRP54  -0.19
        CCT2   0.10
       LYRM4  -0.02
        MDN1   0.32
       POLD1   0.30
     RABGGTB  -0.07
      ATP2A2  -0.01
       TTC27   0.24
      PSMD13   0.15
       EIF3E   0.21
       ALG14  -0.08
         NVL   0.23
     HTATSF1   0.02
       ZRSR2   0.06
         OGT   0.02
        RPN1  -0.02
       SF3B2   0.29
        DDB1   0.23
        NUS1  -0.04
       RBMX2  -0.19
      MMS22L   0.26
      TIMM8A   0.15
      ZNF830   0.13
        ALG2   0.09
     RNF113A   0.28
       INTS1   0.01
       PSMG4  -0.16
       DDOST  -0.20
      SNAPC1   0.04
       NDC80   0.05
        NKAP   0.27
      EIF2B2  -0.17
       INTS4   0.03
      EXOSC8   0.07
        TUT1   0.20
       RPAP1  -0.01
       KIF23   0.26
     PLA2G10  -0.09
       SPC25   0.31
       PSMA7  -0.00
    ANKRD36B  -0.07
       PSMC6   0.27
     PRPF38B   0.06
        KRI1   0.22
      SNAPC2   0.05
        POP5   0.20
       GAPDH  -0.10
       SF3A3   0.34
     TRAPPC1   0.15
    Here is a strategy to follow: Update your priors appropriately and choose genes that gave you hits. Also, be sure to explore by including some genes that could give hits.
    Please propose 5 different yet valid gene names as per the HGNC nomenclature you want to explore next. Note that I will choose unexplored genes closest to your predicted genes to form the predictions. Your response should exactly follow the format:
    **Reflection: Thoughts on previous results and next steps.
    **Research Plan: The full high level research plan, with current status and reasoning behind each proposed approach. It should be at most 5 sentences.
    **Solution:
    ## <Gene 1>
    ## <Gene 2>
    ...
    ## <Gene 5>
    Each gene in the solution should only be the gene name in the HGNC nomenclature.
    DO NOT ADD ANY COMMENTS IN THE SOLUTION OR AFTER THE SOLUTION.
  )
    ----END PROMPT----
    ----BEGIN OUTPUT----
    **Reflection: Thoughts on previous results and next steps.**
From the previous rounds of experimentation, we have identified several genes that regulate the production of Interleukin-2 (IL-2). The top-scoring genes from previous rounds include WDR5, SF3A2, SNRNP200, RPL27, RPL3, RPL15, RPL10A, RPL8, MAK16, RPL19, CPSF4, SNRPF, RPS27A, RPL36, RPS24, RPL37, RPL32, RPL18, RPS8, RPLP2, RPS13, RPL23A, RPL2, RPL26, RPL35, RPS17, RPL14, RPS7, RPS29, RPS25, RPS3, RPS4X, RPS18, RPS16, RPS15, RPS14, RPS13, RPS12, RPS11, RPS10, RPS9, RPS8, RPS7, RPS6, RPS5, RPS4, RPS3, RPS2, RPS1, RPS0A, RPS0L1, RPS0L2, RPS0L3, RPS0L4, RPS0L5, RPL30, RPL29, RPL28, RPL27, RPL26, RPL25, RPL24, RPL23, RPL22, RPL21, RPL20, RPL19, RPL18, RPL17, RPL16, RPL15, RPL14, RPL13, RPL12, RPL11, RPL10, RPL9, RPL8, RPL7, RPL6, RPL5, RPL4, RPL3, RPL2, RPL1, RPL0, RPS27, RPS26, RPS25, RPS24, RPS23, RPS22, RPS21, RPS20, RPS19, RPS18, RPS17, RPS16, RPS15, RPS14, RPS13, RPS12, RPS11, RPS10, RPS9, RPS8, RPS7, RPS6, RPS5, RPS4, RPS3, RPS2, RPS1, RPS0A, RPS0L1, RPS0L2, RPS0L3, RPS0L4, RPS0L5, RPS0L6, RPS0L7, RPS0L8, RPS0L9, RPS0L10, RPS0L11, RPS0L12, RPS0L13, RPS0L14, RPS0L15, RPS0L16, RPS0L17, RPS0L18, RPS0L19, RPS0L20, RPS0L21, RPS0L22, RPS0L23, RPS0L24, RPS0L25, RPS0L26, RPS0L27, RPS0L28, RPS0L29, RPS0L30, RPS0L31, RPS0L32, RPS0L33, RPS0L34, RPS0L35, RPS0L36, RPS0L37, RPS0L38, RPS0L39, RPS0L40, RPS0L41, RPS0L42, RPS0L43, RPS0L44, RPS0L45, RPS0L46, RPS0L47, RPS0L48, RPS0L49, RPS0L50, RPS0L51, RPS0L52, RPS0L53, RPS0L54, RPS0L55, RPS0L56, RPS0L57, RPS0L58, RPS0L59, RPS0L60, RPS0L61, RPS0L62, RPS0L63, RPS0L64, RPS0L65, RPS0L66, RPS0L67, RPS0L68, RPS0L69, RPS0L70, RPS0L71, RPS0L72, RPS0L73, RPS0L74, RPS0L75, RPS0L76, RPS0L77, RPS0L78, RPS0L79, RPS0L80, RPS0L81, RPS0L82, RPS0L83, RPS0L84, RPS0L85, RPS0L86, RPS0L87, RPS0L88, RPS0L89, RPS0L90, RPS0L91, RPS0L92, RPS0L93, RPS0L94, RPS0L95, RPS0L96, RPS0L97, RPS0L98, RPS0L99, RPS0L100, RPS0L101, RPS0L102, RPS0L103, RPS0L104, RPS0L105, RPS0L106, RPS0L107, RPS0L108, RPS0L109, RPS0L110, RPS0L111, RPS0L112, RPS0L113, RPS0L114, RPS0L115, RPS0L116, RPS0L117, RPS0L118, RPS0L119, RPS0L120, RPS0L121, RPS0L122, RPS0L123, RPS0L124, RPS0L125, RPS0L126, RPS0L127, RPS0L128. These genes are highly enriched in ribosomal proteins and other proteins involved in RNA processing and translation. We will update our priors to reflect the importance of these gene families in IL-2 regulation. We will also explore genes that are closely related to these families, such as other ribosomal proteins and RNA processing genes.

**Research Plan:**
We will choose genes that are closely related to the highly scoring genes from previous rounds. Our goal is to identify new genes that regulate IL-2 production and to refine our understanding of the underlying biological pathways.

**Solution:**
## RPL38
## RPL31
## RPL18A
## SNRNP70
## RPS27
    ----END OUTPUT----
SECURED TOTAL 169 HITS
\end{lstlisting}

\end{document}